\documentclass[11pt]{article}

\usepackage[utf8]{inputenc}
\usepackage[T1]{fontenc}
\usepackage{lmodern}
\usepackage{textcomp}

\usepackage{indentfirst}

\usepackage{amsmath,amssymb}
\usepackage{graphicx}
\usepackage{booktabs,longtable,array,multirow}
\usepackage{calc} 
\usepackage{caption}

\usepackage{etoolbox}
\makeatletter
\patchcmd\longtable{\par}{\if@noskipsec\mbox{}\fi\par}{}{}
\makeatother

\usepackage{geometry}
\usepackage{xurl}
\usepackage[hidelinks]{hyperref}

\usepackage{footnote}
\makesavenoteenv{longtable}

\makeatletter
\def\maxwidth{\ifdim\Gin@nat@width>\linewidth\linewidth\else\Gin@nat@width\fi}
\def\maxheight{\ifdim\Gin@nat@height>\textheight\textheight\else\Gin@nat@height\fi}
\makeatother
\setkeys{Gin}{width=\maxwidth,height=\maxheight,keepaspectratio}

\begin{document}

\title{\LARGE \textbf{Spinal Line Detection for Posture Evaluation through Training-free 3D Human Body Reconstruction with 2D Depth Images}}
\author{
  \textbf{Sehyun Kim}\textsuperscript{1}\thanks{Equal contribution} \quad
  \textbf{Hye Jun Lee}\textsuperscript{1}\footnotemark[1] \quad
  \textbf{Jiwoo Lee}\textsuperscript{1}\footnotemark[1] \quad
  \textbf{Changgyun Kim}\textsuperscript{2}\thanks{Corresponding author} \quad
  \textbf{Taemin Lee}\textsuperscript{2}\footnotemark[2]
  \\[6pt]
  \normalsize\textsuperscript{1}Department of Artificial Intelligence and Software, Kangwon National University, Republic of Korea\\
  \small\href{mailto:about7086@kangwon.ac.kr}{about7086@kangwon.ac.kr},\,
  \href{mailto:lovo6027@kangwon.ac.kr}{lovo6027@kangwon.ac.kr},\,
  \href{mailto:jiwo0723@kangwon.ac.kr}{jiwo0723@kangwon.ac.kr}
  \\[4pt]
  \normalsize\textsuperscript{2}Department of Electronic and AI System Engineering, Kangwon National University, Republic of Korea\\
  \small\href{mailto:kevinlee@kangwon.ac.kr}{kevinlee@kangwon.ac.kr},\,
  \href{mailto:tiockdrbs@kangwon.ac.kr}{tiockdrbs@kangwon.ac.kr}
}
\date{}
\maketitle

\begin{abstract}
 The spinal angle is an important indicator of body balance. It is
important to restore the 3D shape of the human body and estimate the
spine center line. Existing multi-image-based body restoration methods
require expensive equipment and complex procedures, and single
image-based body restoration methods have limitations in that it is
difficult to accurately estimate the internal structure such as the
spine center line due to occlusion and viewpoint limitation. This study
proposes a method to compensate for the shortcomings of the
multi-image-based method and to solve the limitations of the
single-image method. We propose a 3D body posture analysis system that
integrates depth images from four directions to restore a 3D human model
and automatically estimate the spine center line. Through hierarchical
matching of global and fine registration, restoration to noise and
occlusion is performed. Also, the Adaptive Vertex Reduction is applied
to maintain the resolution and shape reliability of the mesh, and the
accuracy and stability of spinal angle estimation are simultaneously
secured by using the Level of Detail ensemble. The proposed method
achieves high-precision 3D spine registration estimation without relying
on training data or complex neural network models, and the verification
confirms the improvement of matching quality.
\end{abstract}

\section{Introduction}

The posture and balance of a person serve as important indicators of
overall physical health, extending beyond simple movement patterns.
Various studies have shown that they are directly related to pain,
musculoskeletal disorders, fall risk, and quality of life. Kumar et al.
[1] noted that the registration and curvature of the spine,
including the cervical vertebrae, are closely associated with
musculoskeletal disorders. Additionally, Ishikawa et al. [2] found
that spinal mobility and sagittal registration are directly connected to
the quality of life and fall risk in older adults. Du et al. [3] and
Hussein et al. [4] also emphasized that analyzing the spine is a
crucial measure for understanding physical health. Furthermore, Smythe
et al. [5] pointed out that while the concept of "good posture" is
ambiguous, the evaluation of posture remains significant. These
preceding studies indicate that estimating the spinal centerline and
analyzing human posture based on it is an important task in fields like
rehabilitation medicine, ergonomics, and healthcare.

However, because the position of the spine corresponds to internal human
structures, accurately determining the central axis using only typical
2D images is challenging. To overcome this limitation, various 3D human
reconstruction techniques have been proposed. Methods based on multiple
RGB cameras, as noted by Nogueira et al. [6], offer high accuracy
but require expensive equipment and complex calibration procedures,
limiting their use in general environments. In contrast, approaches
using depth images have garnered attention for being relatively
low-cost, leading to numerous studies [7-10]. Research utilizing
single-point depth images has been actively conducted [11,12].
However, as highlighted in Zhang et al.'s research [13], there
remains a limitation in reliably estimating the entire human shape or
internal central structures like the spinal axis, due to issues of
occlusion from the frontal and rear perspectives.

To address these issues, this study proposes a system for reconstructing
a 3D model of the human body by combining Depth Maps obtained from four
directions: front, back, left, and right, and automatically estimating
the spinal centerline from it. The proposed system converts the depth
data from each direction into Point Clouds and undergoes two stages:
global registration based on RANSAC (Random Sample Consensus) and FPFH
(Fast Point Feature Histograms), followed by fine registration using the
ICP (Iterative Closest Point) algorithm to create an integrated 3D human
model. The generated model is refined through a Vertex Reduction process
to remove noise and correct the human shape, ensuring a stable human
model. A morphological analysis is then performed on this 3D model to
estimate the position of the spine.

The core of this research is to present a method for inferring the
central structures of the human body solely through pure geometric
registration and morphological analysis, without reliance on complex
training data or deep learning models. This approach avoids issues such
as data sparsity and privacy regulations prevalent in medical imaging
contexts, while ensuring that each stage of the algorithm is clear and
interpretable, facilitating verification. Consequently, this study
presents a 3D human analysis pipeline that can be utilized in both
medical and non-medical environments, possessing the potential for
expansion into various fields such as healthcare, rehabilitation
medicine, posture correction, ergonomic design, and digital human
modeling.

\section{Background Researches}

3D human body reconstruction plays a vital role in various applications,
including medical diagnosis, body shape analysis, and virtual fitting.
Particularly, 3D modeling based on medical imaging provides essential
information for accurately understanding a patient's body shape, which
is crucial for diagnosis and treatment planning. Recent research can be
categorized based on methods that utilize single RGB images, multi-view
images, and depth information. Depth map-based 3D reconstruction offers
fine geometric information that can be difficult to obtain using only an
RGB camera, significantly enhancing the accuracy of shape restoration.
In the context of medical imaging, depth information obtained from CT
scans or depth sensors enables the accurate representation of body
shapes while maintaining anatomical consistency.

The fusion of multi-view depth maps effectively addresses the occlusion
problem of single views, allowing for the generation of complete 3D
models. Approaches that use four-directional depth maps integrate
information from the front, back, left, and right to create a fully
comprehensive human model. Point Cloud registration is a crucial step in
3D reconstruction, involving the integration of Point Cloud data
acquired from different viewpoints into a single, consistent coordinate
system. Traditional registration methods are generally composed of two
main stages: feature-based global registration and iterative refinement,
which help ensure accurate registration of the data.

\subsection{Point Cloud Registration}

\subsubsection{Feature-based Global Registration}

RANSAC (Random Sample Consensus) [14] has established itself as a
standard method for estimating robust transformation matrices from data
containing outliers. RANSAC removes outliers through random sampling and
performs initial registration using only inliers, demonstrating
relatively stable performance even in environments with noise and
partial overlap. However, it may fail to converge in extreme cases where
the outlier ratio exceeds 50\%, and its computational cost can
significantly increase with the number of iterations.

FPFH (Fast Point Feature Histograms) [15] is a geometric feature
descriptor that quantifies the normals, curvature, and surface curvature
patterns around each point into a 3-dimensional histogram. FPFH enables
more robust correspondence point detection compared to simple
distance-based matching; however, it is sensitive to noise and shows
reduced discriminative power in flat areas. Research by Zheng [16]
has reported a decrease in accuracy of FPFH in the fusion of
multi-source Point Clouds with noise and local distortions, proposing a
CNN-based virtual correspondence generation method to overcome this.
Moreover, in flat surfaces, the small curvature variation reduces the
discriminative capability of the FPFH histogram, leading to mismatches
in correspondence points in relatively flat regions of the human body,
such as the back, chest, and abdomen.

Recently, Chu et al. [17] proposed a global alignment and stitching
method for multi-view point clouds using 2D SIFT (Scale-Invariant
Feature Transform) based feature points. This approach extracts SIFT
features from multiple camera images and matches them to 3D point cloud
coordinates, demonstrating robust alignment results even in repetitive
structures or complex scenes. However, there is a limitation that it can
only be effectively applied when sufficient texture or repetitive
patterns are present.

Additionally, neural implicit representation-based registration methods
have emerged. NeRF (Neural Radiance Fields) and NICE-SLAM [18] can
model input data as continuous functions, maintaining structural
consistency of the entire scene in high-dimensional embedding space
while minimizing local feature loss. This neural network-based approach
helps overcome the limitations of traditional discrete descriptors,
contributing to fast real-time processing and high-quality 3D
reconstruction, although it entails high demands for large-scale
parameters and dependence on training data.

Among these methods, RANSAC and FPFH are robust against noise and
computationally efficient, enabling effective feature-based matching
during global initial registration. In this research, we leverage these
characteristics to first perform global registration using RANSAC and
FPFH, followed by local refinement using the ICP (Iterative Closest
Point) algorithm. This procedure significantly enhances the precision of
the final 3D model while ensuring both the stability of the initial
registration and the accuracy of local optimization.

\subsubsection{Precision Refinement: Limitations of Point-to-Point and}
Surface-based Approaches

The ICP (Iterative Closest Point) algorithm [19] is a representative
method for fine registration of Point Clouds. Traditional Point-to-Point
ICP iteratively finds the nearest points between two Point Clouds and
computes a rigid transformation to minimize the sum of squared distances
between the points. However, this approach has critical limitations,
including slow convergence speed, susceptibility to getting stuck in
local optima, and the potential to converge in entirely different
directions when the initial registration is inaccurate.

Go-ICP [20] stated that "ICP can only guarantee convergence to local
minima due to its iterative nature," and introduced
Branch-and-Bound-based global optimization to address this issue. Low et
al. pointed out that ICP often gets trapped in local minima close to the
optimal solution and proposed multi-resolution surface smoothing as a
potential solution. However, these methods also significantly increase
computational costs, making them unsuitable for real-time applications
such as in healthcare.

To overcome these problems, Point-to-Plane ICP was proposed, which
achieves significantly faster convergence rates and higher accuracy by
minimizing the perpendicular distance between points and tangential
planes. Rusinkiewicz et al. [21] demonstrated that the
Point-to-Plane approach performs excellently on data rich in surface
information, with convergence speed improving several times when normal
vectors can be reliably estimated. Furthermore, Plane-to-Plane ICP
[22] utilizes plane information from both source and target Point
Clouds to perform surface-normal guided registration. This approach
considers the local planar structure of both Point Clouds
simultaneously, providing a broader geometric context than point-wise
correspondences and increasing the potential to avoid local optima.

Förstner [23] theoretically demonstrated that efficient and accurate
motion estimation is possible from plane-based correspondences. Chen et
al. [24] empirically showed that plane-based alignment can achieve
high accuracy and robustness in structured environments, such as
building facades, as well as in shapes exhibiting partial plane
characteristics, like human surfaces, using real point cloud data. In
the same study, they utilized the PLADE (Plane-based Descriptor) to
demonstrate that plane information is effective in 3D feature
representation, proving that it enables stable alignment in relatively
flat areas of the human body, such as the back, chest, abdomen, and
thighs.

\subsubsection{Applying Vertex Reduction after ICP Registration}

Vertex reduction of Point Clouds is typically performed during the
preprocessing stage for the purposes of operational efficiency and data
light weighting. However, recent studies have increasingly applied
vertex reduction in the post-processing stage after registering
(aligning) Point Clouds from multiple viewpoints. Efficiently
simplifying the data after registration allows for the effective removal
of redundant areas and unnecessary vertices, while preserving geometric
features (such as curvature and edges) and maintaining accuracy.

Lawin et al. [25] employed a density-adaptive point set registration
technique, conducting ICP-based alignment on multi-view data with
differing densities, followed by post-processing downsampling to correct
observed density differences in the aligned results. In terms of feature
preservation, Potamias et al. [26] proposed a vertex reduction
method that selects salient points based on learned features of the
point cloud's structural characteristics (such as edges and curvature),
demonstrating that high simplification rates can still achieve excellent
shape preservation and geometric consistency. Additionally, Xueli Xu et
al. [27] reported a case where a feature-preserving mesh
simplification technique was applied in the post-processing stage in
densely aligned 3D microscopy data, maintaining the quality of the final
model.

Post-processing vertex reduction, especially in the context of
integrated Point Clouds after ICP registration, serves as a tool to
flexibly adjust the trade-off between data quality and operational
efficiency according to needs. For example, in medical settings, regions
that require high precision (e.g., areas where surface curvature changes
rapidly) can minimize reduction, while flatter areas can be aggressively
simplified. This strategy reduces the overall data volume while
effectively preserving essential information.

\subsection{Deep Learning-based 3D Reconstruction: Fundamental Limitations in}
the Healthcare Data

Deep learning-based 3D reconstruction methods have been rapidly
evolving. PointNet-based matching networks(PointNetLK, DCP, and PCRNet)
and Transformer-based multi-view fusion models are representative
examples, demonstrating impressive performance through training on large
datasets.

Notably, PointNetLK [28] combines the PointNet architecture with
Lucas-Kanade optimization to implement robust rigid registration between
two 3D point sets. The advantages of PointNetLK include direct and fast
registration without using RANSAC, as well as excellent noise robustness
inherent in the PointNet framework. However, its optimization
performance may degrade in clinical data or highly noisy environments,
and it has a significant dependency on large training datasets.

DCP (Deep Closest Point) [29] incorporates an Attention mechanism to
enable fine correspondence generation and transformation estimation
between two Point Clouds. This model guarantees robust performance in
complex transformation situations and is adaptable to various
environments, but it also suffers from a complex network structure and
increased training and inference times. Additionally, like PointNetLK,
DCP requires sufficient annotated large-scale data to achieve optimal
performance.

PCRNet [30] operates by incrementally increasing registration
precision through a recurrent neural network (RNN) structure integrated
with PointNet. The main advantage of PCRNet is that it exhibits rapid
convergence and robust registration performance even with a relatively
small network. However, it may be vulnerable to convergence stability
and errors in cases of high data complexity or significant noise.

As highlighted, while deep learning-based 3D reconstruction demonstrates
excellent registration performance in large general datasets, it faces
fundamental limitations in the medical data environment.

First, there is the issue of data scarcity. Due to privacy regulations
and high collection costs, it is extremely difficult to secure the
substantial amounts needed for large-scale training in medical data. A
comprehensive review by Chen et al. [31] identified data scarcity
and labeling costs as significant obstacles to deep learning in medical
imaging, especially pointing out that annotation work is even more
complex and time-consuming for 3D data. The research by Xiao et al.
[32] noted the inaccuracies and noise problems in depth-based motion
capture data, emphasizing the need for additional training data to
correct these issues, which is challenging to obtain in medical
settings.

Second, there is the challenge of shape diversity and generalization.
Even for the same patient, subtle differences in posture, body shape,
and capturing conditions can significantly hinder the generalization of
the learned models. In medical applications, constructing training
datasets that cover various body types (e.g., obese, muscular,
underweight), ages (e.g., pediatric, adult, elderly), and posture
variations is virtually impossible. Research by Guan et al. [33]
systematically analyzed the domain adaptation issues of medical deep
learning models, reporting a sharp decline in performance on new patient
data with distributions different from the training data.

Third, ethical and security constraints pose significant challenges.
There is a risk of patient data being leaked during the deep learning
model training process, as well as concerns about the ability to trace
back original data from the trained models. Furthermore, the black-box
nature of these models makes it difficult to provide the
interpretability and decision-making transparency required in medical
practice, which can pose significant obstacles in the regulatory
approval process for medical devices.

\subsection{The Contributions and Approach methods of our paper}

This research aims to construct an accurate 3D human model from
four-directional medical depth maps, featuring structural
distinctiveness compared to existing studies. We propose a hierarchical
registration pipeline using RANSAC-FPFH-Plane ICP and adopt
Plane-to-Plane ICP-based precision registration instead of the
conventional Point-to-Point ICP method. The Plane-to-Plane approach is
particularly suitable for medical depth data, as it utilizes surface
normal information to achieve faster convergence rates than
Point-to-Point methods, thereby reducing processing time required in
clinical workflows. Additionally, plane information provides a broader
geometric context than point information, decreasing the likelihood of
getting trapped in local optima, and allows for quick and stable
convergence to the global optimum when the initial registration obtained
from the four-directional fixed capture is relatively good.

Especially for body parts with strong planar characteristics, such as
the back, chest, abdomen, and thighs, Plane-to-Plane correspondences are
significantly more stable and accurate than Point-to-Point
correspondences. By explicitly considering the surface normal
consistency between viewpoints in the four-directional fixed capture
environment, the registration quality across viewpoints can be further
improved. Moreover, this methodology operates purely based on geometric
principles without relying on large-scale training data or external
data, fundamentally circumventing the limitations of medical data, such
as data scarcity, ethical constraints, and difficulties in
generalization. Each step of the algorithm is clear and interpretable,
ensuring the verifiability and reliability required in medical device
regulatory approval processes.

Through this approach, this research presents a clinically applicable 3D
human reconstruction method that simultaneously achieves accuracy,
stability, interpretability, and practicality required in medical
environments without the need for complex deep learning models.

\section{Proposed Method}

\subsection{System Overview}

The 3D human posture analysis system proposed in this study operates on depth maps generated under controlled conditions and consists of a multi-stage pipeline that extracts high-quality 3D meshes and spinal skeletal information from depth images. The depth maps used in this study were generated using data obtained from the SizeKorea database, which is a nationally managed anthropometric survey that collects and pro-vides standardized high-resolution 3D human body models. [34]

The system first receives depth images from four directions (front, left, right, and back) to generate 3D Point Clouds for each view. In this process, a mask image is applied for background removal, and points are generated according to the unique coordinate sys-tems of each view. Since the Point Clouds generated from multiple views possess different local coordinate systems, a registration process is essential to integrate them into a single global coordinate system. This system adopts a two-step registration strategy. The first step is global registration, which combines the RANSAC (Random Sample Consensus) algorithm with the FPFH (Fast Point Feature Histograms) feature descriptor. The second step utilizes the ICP (Iterative Closest Point) algorithm for fine registration. After the global initialization, we incorporate a quality-driven feedback mechanism that quantitatively assesses geometric consistency among the aligned views and automatically re-runs the alignment with adjusted parameters when the quality is insufficient. After initial registra-tion, the Point-to-Plane ICP algorithm is applied to achieve fine registration. The registra-tion process proceeds in the order of side views (left and right) followed by the back view, structured to minimize cumulative errors by aligning in the order of higher view overlap.

Once the registration is complete, the Point Clouds are merged into a
unified Point Cloud, which is then optimized into a six-level LOD (Level
of Detail) model through a Quadric Error Metrics-based vertex reduction
algorithm. An ensemble technique is applied to independently extract
spinal skeletal information from each LOD model. In this process,
AI-based landmark detection is combined with anatomical proportions to
estimate 17 key joint points and compute spinal angles. The spinal
skeletal information predicted from the six LOD models is integrated
using a voting method, effectively eliminating noise due to resolution
differences by selecting the most frequently occurring coordinate for
each joint point. This multi-resolution-based ensemble strategy enhances
the robustness and accuracy of spinal skeletal predictions compared to a single model.

\includegraphics[width=7.14167in,height=3.06736in]{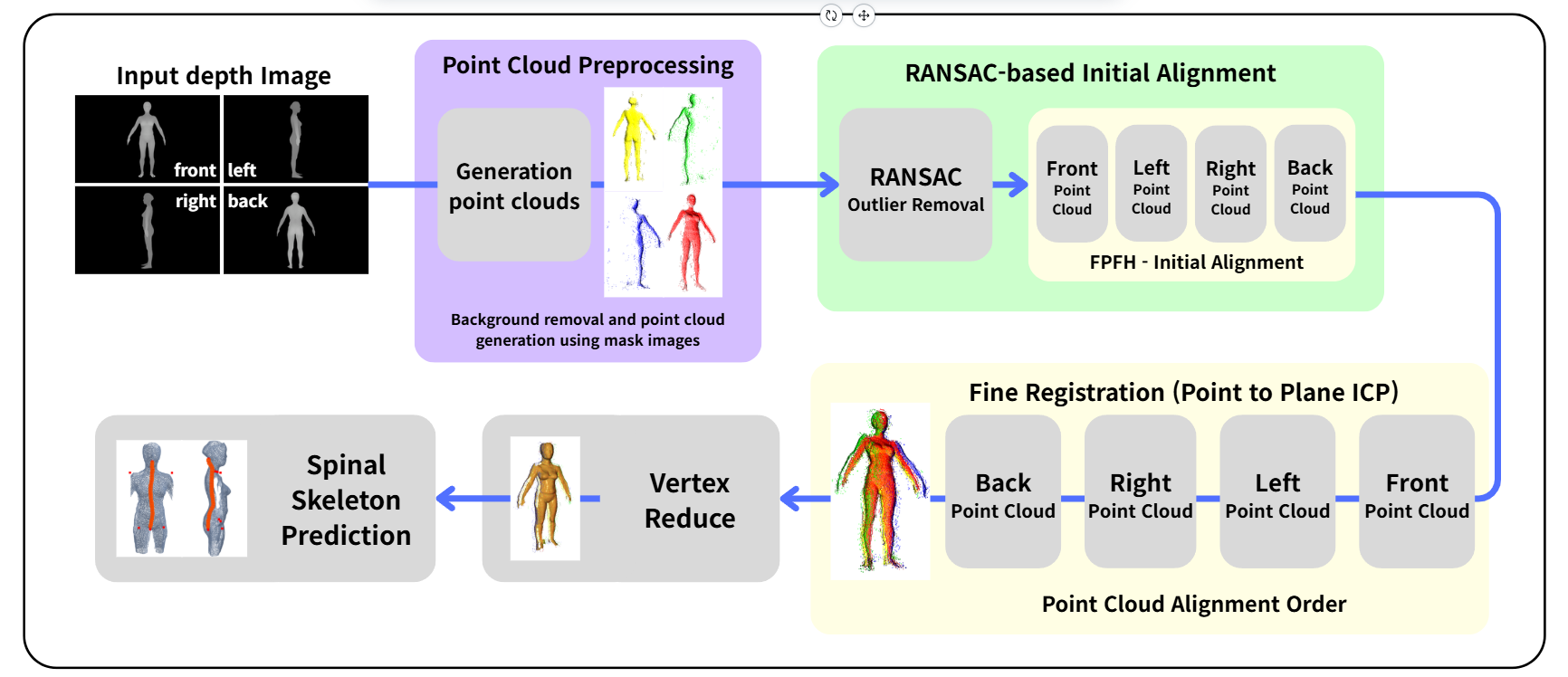}
\captionof{figure}{System flow of proposed method}\label{fig:1}

\subsection{Preprocessing: Depth Map Analysis and Point Cloud Generation}

This system receives depth images from four directions (front, left,
right, and back) to generate 3D Point Clouds. The depth images are
provided in bmp format as grayscale images, where the brightness value
of each pixel represents distance information from the camera.

As can be seen in the No Mask Preprocessing example in Figure 2,
directly converting the Depth Image to a point cloud resulted in
excessive measurement noise, leading to significant distortion along the
edges of the Point Cloud. To address this issue, this study introduced a
mask-based preprocessing step to finely extract the foreground regions
from the depth image. This allows for the proactive removal of
unnecessary background points and sensor-induced noise, facilitating
stable Point Cloud reconstruction.

First, the input depth map is represented with integer values ranging
from 0 to 255, but for numerical stability in subsequent processing, it
is normalized to floating-point values in the range of 0.0 to 1.0. An
adaptive mask generation method based on double thresholding is applied
to efficiently separate the foreground (the human body) from the
background. The lower threshold of 0.2 removes sensor noise and overly
close areas, while the upper threshold of 0.95 filters out regions
corresponding to the background and depth measurement limits.

The generated binary mask undergoes morphological opening and closing
operations using elliptical structural elements, which helps eliminate
residual noise and naturally refine the outline of the human body. The
depth map with the refined mask is then transformed into a 3D Point
Cloud using an inverse projection algorithm based on the pinhole camera
model. A KD-Tree-based hybrid neighbor search is conducted to extract up
to 30 neighboring points within a 5 mm radius around each point, and
Principal Component Analysis (PCA) is applied to this local patch to
estimate the normal vectors. The eigenvector corresponding to the
smallest eigenvalue is chosen as the normal direction, and this
high-quality normal information significantly enhances the precision of
the subsequent Point-to-Plane ICP registration process.

\begin{center}
\includegraphics[width=5.42672in,height=3.8674in]{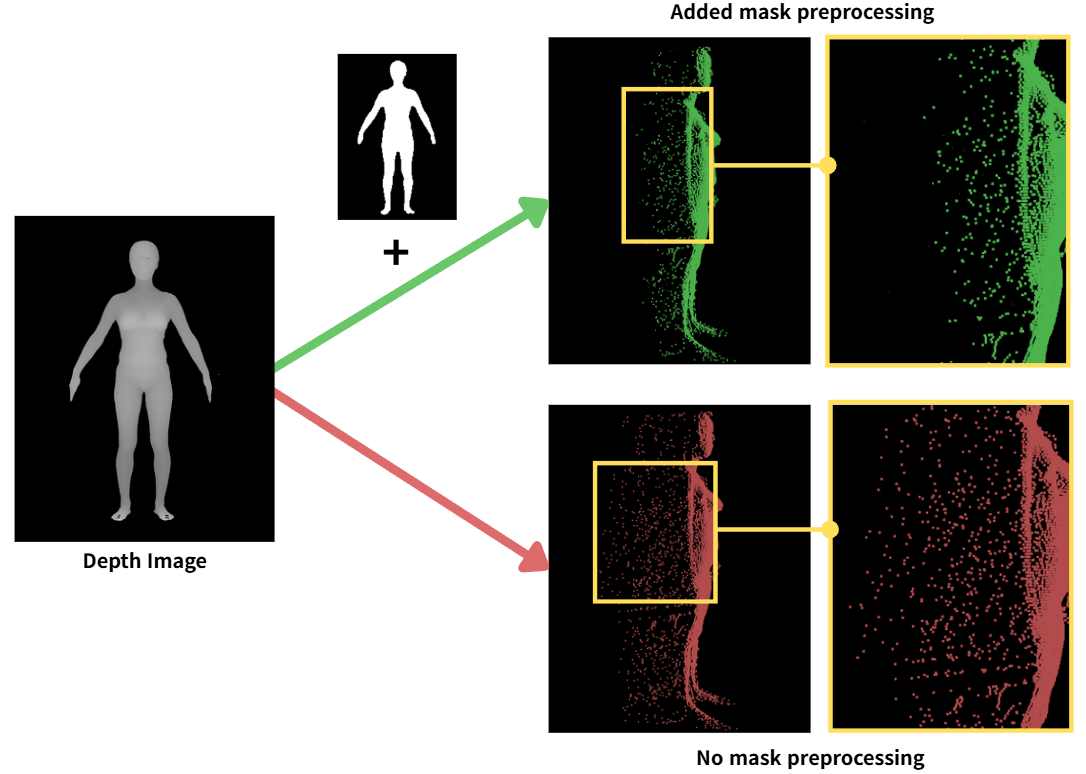}
\end{center}

\captionof{figure}{Difference of extracted Point Cloud with mask/no mask.}\label{fig:2}

\subsection{FPFH Feature Descriptor Extraction}

The ICP (Iterative Closest Point) algorithm updates the nearest
correspondences between two Point Clouds iteratively to optimize the
transformation (rotation/translation). Consequently, due to the
convergence characteristics of the algorithm, the initial pose is a key
factor determining the overall registration quality. ICP is structured
such that its objective function is nonlinear and easily converges to
local minima. If the initial registration is inaccurate or if the
relative distance between the two Point Clouds is significant, it
continuously matches incorrect correspondences, leading to the
accumulation of misalignments. Especially in structures with significant
curvature changes, such as the human body, an inadequate initial pose
can greatly distort the relative positions of arms, legs, and the torso,
making it virtually impossible to recover in subsequent optimization
iterations.

For this reason, fine registration using ICP must be conducted under
conditions where a stable initial pose is ensured. As can be seen in
Figure 3, applying ICP without initial registration can lead to
misalignments that geometrically distort or overlap the overall shape by
getting trapped in local optima.

To prevent this, the current study stabilizes the initial pose using
global registration based on RANSAC-FPFH and then performs fine
registration using Point-to-Plane ICP. To address this, the system
utilizes RANSAC (Random Sample Consensus)-based FPFH (Fast Point Feature
Histograms) feature matching to achieve global initial registration,
followed by fine registration through the Point-to-Plane ICP algorithm.
This combined approach secures global exploration capabilities during
initial registration and achieves high registration accuracy through
local optimization in the fine registration phase.

\begin{longtable}[]{@{}
  >{\raggedright\arraybackslash}p{(\columnwidth - 2\tabcolsep) * \real{0.4873}}
  >{\raggedright\arraybackslash}p{(\columnwidth - 2\tabcolsep) * \real{0.5127}}@{}}
\toprule
\endhead
\includegraphics[width=2.50688in,height=2.09351in]{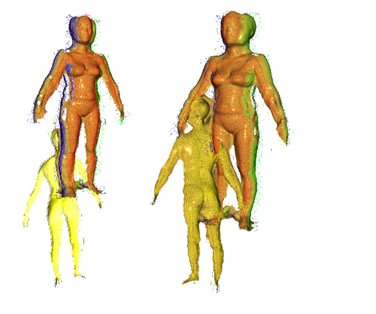}
&
\includegraphics[width=2.46021in,height=2.08685in]{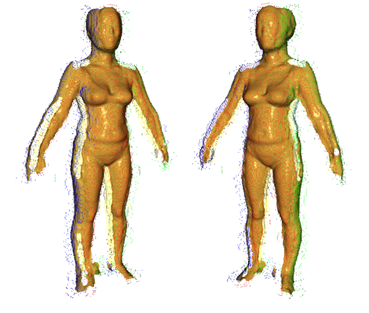} \\
\textbf{(a)} ICP only & \textbf{(b)} RANSAC\_FPFH+ICP \\
\bottomrule
\end{longtable}

\captionof{figure}{Comparison of output quality when adding an initial RANSAC-based Initial Registration (We used Point-to-Plane ICP for
comparison}\label{fig:3}

\subsubsection{Depth Normalization and Mask-Based Point Cloud Generation}

FPFH (Fast Point Feature Histograms) is a structurally robust local
feature descriptor that effectively addresses issues such as partial
observations, curvature discontinuities, and depth sensor noise that
inevitably arise in human-based Point Cloud registration. By applying a
center-point-based single accumulation structure, FPFH eliminates the
excessive computational load of the higher-dimensional descriptor (PFH),
which computes the relationship among all pairs of neighboring points,
while reliably preserving the core geometric information of relative
normal distribution.

The feature compression structure of FPFH effectively suppresses noise
propagation commonly encountered in human data and offers a relatively
uniform expression even in areas where curvature changes and
quasi-planar structures coexist, such as at joints and the torso.
Additionally, its low computational cost allows for near real-time
processing speeds in this pipeline, which requires continuous processing
from four viewpoints (front, back, left, right), providing a practical
advantage. Consequently, FPFH is selected as the fundamental component
of local feature descriptors.

\subsubsection{Hierarchical Registration and Supplementary Design Framework}

The system combines RANSAC--FPFH-based initial registration
(Initialization) with a hierarchical adaptive design that reflects the
viewpoint-specific characteristics of the data and practical operational
constraints. This integration ensures both robustness and efficiency in
Point Cloud registration.

Firstly, to address the varying point density and occlusion patterns
depending on the viewpoint, an adaptive voxel downsampling technique was
applied. Voxel sizes of 5.0 mm for the left and back views and 3.0 mm
for the right view were established. A multi-scale progression approach
was introduced to gradually converge from global registration to fine
registration. Specifically, scale stages of [25.0, 12.0, 6.0] mm for
the left and back views and [10.0, 5.0, 2.5, 1.0] mm for the right
view were designed to enable stable convergence from the global contour
to local details. This multi-scale structure allows for comprehensive
representation of the overall shape in the initial stages while
supporting the refinement of local registration in subsequent steps.

Additionally, the rotation degrees of freedom were constrained based on
the viewpoint to minimize misregistration that can occur during the
registration of symmetrical bodies. A limited small-angle rotation was
permitted for the left and back views to ensure registration stability,
while a broader rotation range was allowed for the right view, where
pose variations during data collection are relatively larger, thereby
expanding the convergence domain.

During the RANSAC phase, a fitness threshold was established, and if the
inlier ratio did not meet a certain level, the process automatically
proceeded to the next scale stage. When sufficient matching quality was
achieved, early termination was implemented to reduce unnecessary
sampling. This conditional progression was designed to minimize
computational load while maintaining registration quality.

To improve robustness against occasional failures of global initialization under oc-clusion, noise, or challenging body poses, we introduce a quality-driven feedback mecha-nism that evaluates the geometric consistency of the initially aligned point clouds before proceeding to downstream steps.

After initial alignment, we compute an alignment quality score based on the mean nearest-neighbor distance between sampled points across each aligned view pair. Specif-ically, for a pair of aligned point clouds, we measure the mean nearest-neighbor distance and convert it into an overlap quality in the range [0, 1] by normalizing with the expected body size; the final score is obtained by averaging the overlap quality over all view pairs.

If the alignment quality score falls below a threshold (set to 0.4 in our experiments), the pipeline automatically triggers an adaptive refinement loop that re-invokes the align-ment module with modified settings (e.g., increased RANSAC iterations or alternative ICP strategies) and re-evaluates the result using the same metric. This loop is fully automated and implemented as modular functions (assess\_alignment\_quality and ap-ply\_adaptive\_refinement), ensuring that only geometrically consistent reconstructions are passed to mesh generation and LOD-ensemble skeleton analysis.

\subsubsection{Importance of Registration Order}

The registration order significantly influences the overall registration
quality based on the viewpoint characteristics and shape overlap of each
view. In this study, an appropriate registration sequence was
established for the multiple input Point Clouds. As shown in Figure 4,
applying the correct order allows the four viewpoints to be stably
aligned into a single coherent shape; however, if the order is reversed,
the relative positions of the front, side, and back Point Clouds become
distorted, leading to serious mismatches or overlapping errors. This
occurs because errors from the initial registration phase propagate to
subsequent views, and particularly in human data with many partial
observations, the impact of order selection becomes even more
pronounced.

To minimize this error propagation, this study adopted a gradual
registration procedure of front, left/right side, and back. The front
view was prioritized for several reasons: it most clearly represents the
center axis of the human body and provides stable structural reference
points that define bilateral symmetry, such as shoulder width, hip
width, and thoracic contours. Additionally, the front view tends to have
relatively superior sensor field of view and observation quality, making
it the most suitable reference.

Subsequently, the left and right sides are sequentially aligned to the
previously established front-based target to enhance the transverse
cross-section structure of the human body, and finally, the back Point
Cloud is aligned to complete the overall shape. This order is designed
to perform registration starting from the areas with the highest overlap
between views to suppress cumulative errors, ensuring that the
relatively lower overlapping back view is minimally affected by initial
errors.

\begin{longtable}[]{@{}
  >{\raggedright\arraybackslash}p{(\columnwidth - 6\tabcolsep) * \real{0.2446}}
  >{\raggedright\arraybackslash}p{(\columnwidth - 6\tabcolsep) * \real{0.2446}}
  >{\raggedright\arraybackslash}p{(\columnwidth - 6\tabcolsep) * \real{0.2446}}
  >{\raggedright\arraybackslash}p{(\columnwidth - 6\tabcolsep) * \real{0.2662}}@{}}
\toprule
\endhead
\includegraphics[width=1.62798in,height=1.73214in]{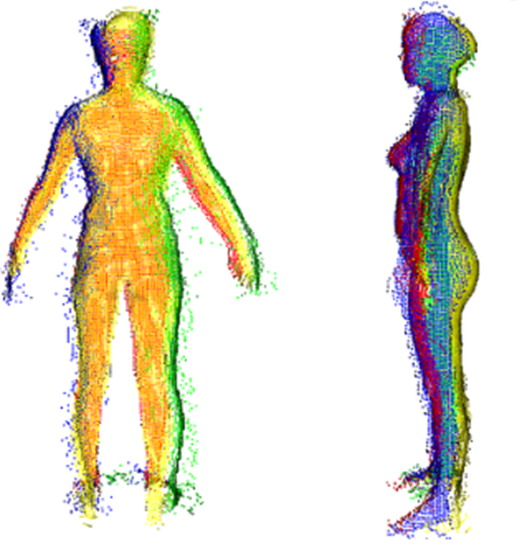}
&
\includegraphics[width=1.62798in,height=1.73214in]{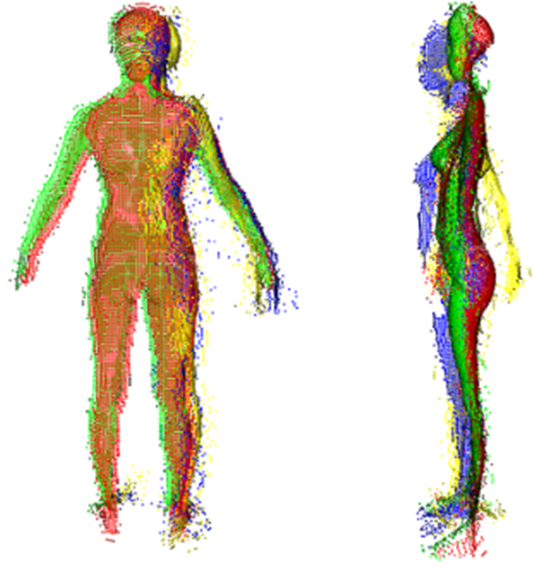}
&
\includegraphics[width=1.62798in,height=1.73214in]{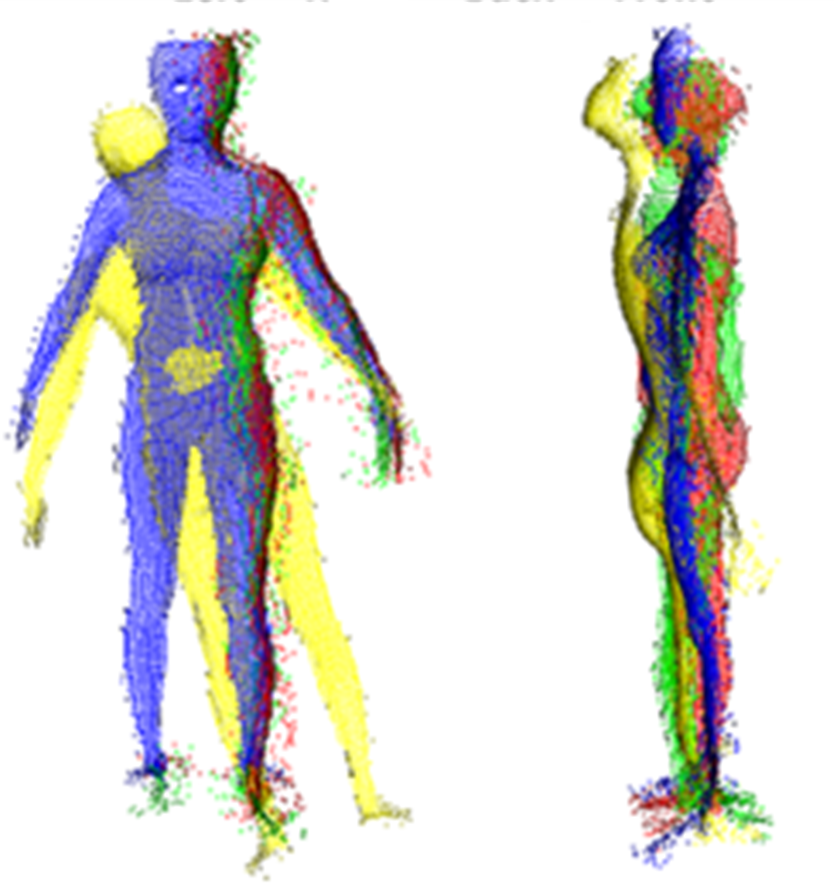}
&
\includegraphics[width=1.62798in,height=1.73214in]{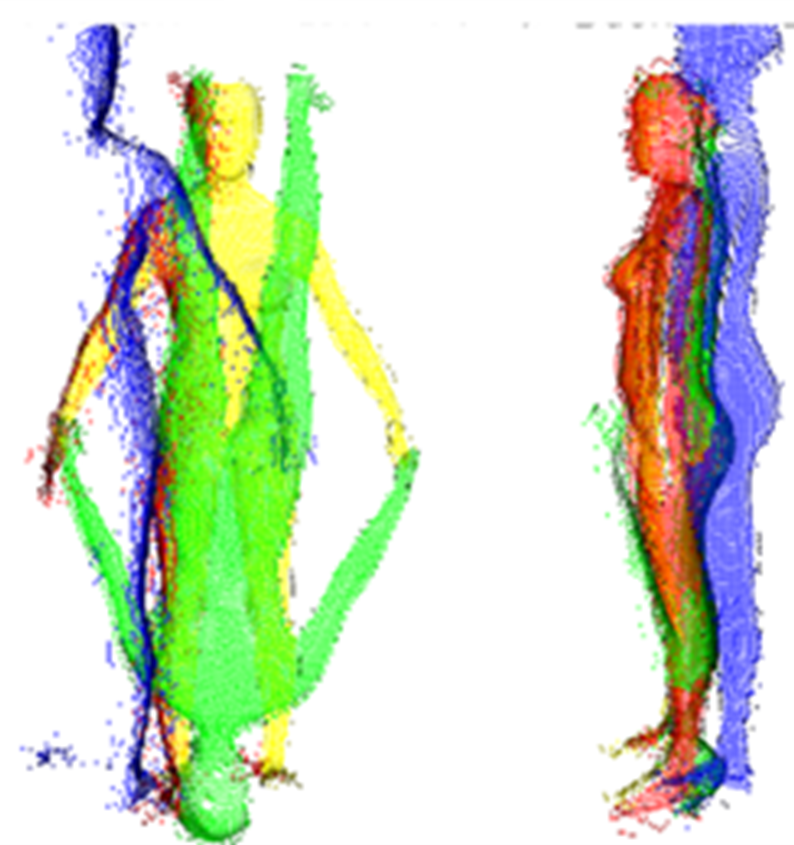} \\
(\textbf{a}) $\mathrm{F} \rightarrow \mathrm{L} \rightarrow \mathrm{R} \rightarrow \mathrm{B}$ &
(\textbf{b}) $\mathrm{B} \rightarrow \mathrm{F} \rightarrow \mathrm{L} \rightarrow \mathrm{R}$ &
(\textbf{c}) $\mathrm{L} \rightarrow \mathrm{R} \rightarrow \mathrm{B} \rightarrow \mathrm{F}$ &
(\textbf{d}) $\mathrm{R} \rightarrow \mathrm{B} \rightarrow \mathrm{L} \rightarrow \mathrm{F}$ \\
\bottomrule
\end{longtable}

\captionof{figure}{Comparison of output results by registration order(F, B, R, and L mean Front, Back, Right, and Left, respectively. (a) is
result of correct order, (b)\textasciitilde(c) are result of wrong
order}

\subsection{Fine Registration (Point to Plane ICP)}

The objective of this stage is to minimize the remaining positional and
orientational discrepancies after the global initial registration
(RANSAC--FPFH) by aligning them to the local surface geometry. Since
human data features extensive quasi-planar or low-curvature regions such
as the thoracic spine, scapula, and pelvis, an accurate fine
registration algorithm is essential.

Figure 5 shows the comparison of the performance of Point-to-Plane ICP
and Point-to-Point ICP by replacing only the cost function under the
same initial global registration and the same correspondence update
procedure. In the full-body registration scenarios, Point-to-Plane ICP
consistently outperformed Point-to-Point ICP in several aspects.

\begin{longtable}[]{@{}
  >{\raggedright\arraybackslash}p{(\columnwidth - 2\tabcolsep) * \real{0.4873}}
  >{\raggedright\arraybackslash}p{(\columnwidth - 2\tabcolsep) * \real{0.5127}}@{}}
\toprule
\endhead
\includegraphics[width=2.46021in,height=2.02017in]{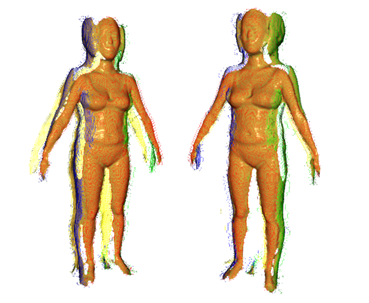}
&
\includegraphics[width=2.46021in,height=2.08685in]{images/image4.png} \\
\textbf{(a)} Point-to-Point ICP & \textbf{(b)} Point-to-Plane ICP \\
\bottomrule
\end{longtable}

\captionof{figure}{Comparison of Point-to-Point ICP and Point-to-Plane ICP Output Quality Performance (After RANSAC-based Initial Registration Process}

 In this stage, RMSE (Root Mean Square Error) and fitness score are used
to quantitatively evaluate the performance of the ICP algorithm. RMSE is
the most widely used metric for measuring alignment accuracy between two
point clouds, quantifying the average distance error between the aligned
point cloud and the reference data. The fitness value ranges from 0 to
1, with higher values indicating that many correspondences between the
two point clouds match. Particularly for partially overlapping point
clouds obtained from medical depth maps, fitness directly reflects the
ratio of valid correspondences contributing to the alignment process,
making it essential for assessing the reliability of the alignment.

Prior studies have comprehensively evaluated alignment results by using
both RMSE and fitness together. RMSE measures the precision of the
alignment, while fitness measures the reliability of the alignment,
allowing for an objective assessment of the overall performance of the
alignment algorithm. This is particularly advantageous in medical
applications, such as this study based on geometric alignment, as it
enables the evaluation of pure geometric performance without the bias of
artificial training data, thereby ensuring clinical reliability.

As shown in Table 1, in the combination of front and left views, both
methods exhibited similar performance (RMSE 4.3246, Fitness 0.6715).
However, in the combination of front and right views, Point-to-Plane ICP
slightly improved the RMSE from 1.6576 to 1.6528. The most notable
difference occurred in the combination of front, left/right, and back
views, where Point-to-Plane ICP reduced the RMSE from 5.2462 to 4.5653,
approximately a 13.0\% decrease, while simultaneously improving the
Fitness value from 0.8024 to 0.8416, an increase of about 4.9\%. This
result clearly illustrates the effectiveness of normal direction
constraints in complex scenarios with significant cumulative errors,
such as back registration.

These findings align with the theoretical characteristics whereby
Point-to-Point ICP is sensitive to slight sliding errors in the surface
tangential direction due to isotropic distance minimization, while
Point-to-Plane ICP utilizes geometric constraints in the surface normal
direction to directly suppress residuals and expand the convergence
domain. Additionally, the dataset exhibits positional noise for
individual points due to the characteristics of the depth sensor;
however, normals are estimated through local averaging, making them
relatively stable. For these reasons, the Fine Registration algorithm in
this process was adopted as Point-to-Plane ICP. The statistical significance of these differences is further validated using paired statistical tests in Section 4.3.

\captionof{table}{Numerical comparison with Point-to-Point ICP and}\label{tab:1}
Point-to-Plane ICP (The lower the value, the better the performance, and
the higher the ICP fitness, the better the performance.)

\begin{longtable}[]{@{}
  >{\raggedright\arraybackslash}p{(\columnwidth - 8\tabcolsep) * \real{0.2776}}
  >{\raggedright\arraybackslash}p{(\columnwidth - 8\tabcolsep) * \real{0.1806}}
  >{\raggedright\arraybackslash}p{(\columnwidth - 8\tabcolsep) * \real{0.1806}}
  >{\raggedright\arraybackslash}p{(\columnwidth - 8\tabcolsep) * \real{0.1806}}
  >{\raggedright\arraybackslash}p{(\columnwidth - 8\tabcolsep) * \real{0.1806}}@{}}
\toprule
\multirow{2}{*}{\begin{minipage}[b]{\linewidth}\raggedright
\end{minipage}} &
\multicolumn{2}{l}{\begin{minipage}[b]{\linewidth}\raggedright
\textbf{Point-to-Point ICP}
\end{minipage}} &
\multicolumn{2}{l}{\begin{minipage}[b]{\linewidth}\raggedright
\textbf{Point-to-Plane ICP}
\end{minipage}} \\
& \begin{minipage}[b]{\linewidth}\raggedright
\textbf{RMSE(mm)}
\end{minipage} & \begin{minipage}[b]{\linewidth}\raggedright
\textbf{ICP fitness}
\end{minipage} & \begin{minipage}[b]{\linewidth}\raggedright
\textbf{RMSE(mm)}
\end{minipage} & \begin{minipage}[b]{\linewidth}\raggedright
\textbf{ICP fitness}
\end{minipage} \\
\midrule
\endhead
Front + Left & 4.3246 & 0.6715 & 4.3246 & 0.6715 \\
Front + Right & 1.6576 & 0.5409 & 1.6528 & 0.5134 \\
4 directions & 5.2462 & 0.8024 & 4.5653 & 0.8416 \\
\bottomrule
\end{longtable}

\clearpage

\subsection{Adaptive Vertex Reduction}

This stage focuses on adjusting computational costs and resolution according
to demands while preserving the shape fidelity of the human mesh after
preprocessing and registration. It minimizes shape distortion and volume
deviation by combining adaptive decimation reflecting curvature, surface
density, and anatomical importance with step-wise normal recalculation
and smoothing. The six Levels of Detail (LOD) generated from the same
original source form a subset hierarchy that ensures consistency in
coordinate systems, boundaries, and normals across resolutions. Figure 6
shows the example of vertex reduction results by LOD.

\includegraphics[width=7in]{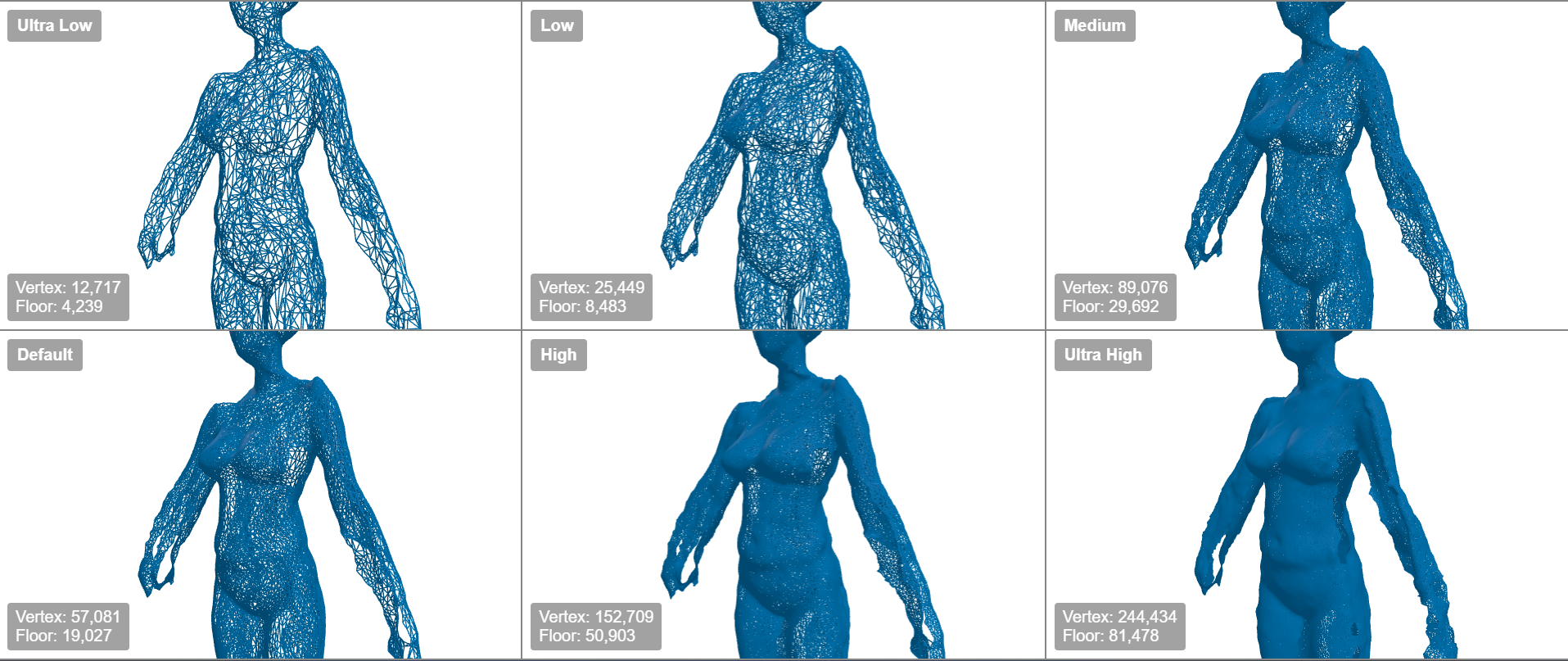}
\captionof{figure}{Examples of Meshed Images by LOD}\label{fig:6}

Adaptive Vertex Reduction removes degenerate, duplicate, and abnormal
elements, recalculates normals, and secures a normalized input mesh,
subsequently reducing the number of vertices step-by-step to meet target
retention rates. The quality-prioritized path utilizes precision
reduction with boundary preservation and low tolerance to minimize
surface distortion and volume deviation. The speed-prioritized path
achieves the same goal by dividing it into five short steps to enhance
throughput. By opting for step division over a single substantial
reduction, cleanup and normal correction are immediately performed at
the end of each step to suppress local collapse, silhouette loss, and
non-manifold remnants. After reduction, lightweight smoothing and
post-processing are applied to alleviate fine jaggedness and noise,
ensuring the stability of sensitive contours and joint silhouettes in
skeleton estimation and measurement. The pipeline operates consistently
in the order of preprocessing, reduction, intermediate cleanup, normal
correction, and smoothing. Table 2 summarizes the reduction rates of the
six LODs produced in this stage

\captionof{table}{The rate of Vertex Reduction depend on LOD. (Reduction}\label{tab:2}
rate was calculated in comparison with the original)

\begin{longtable}[]{@{}
  >{\raggedright\arraybackslash}p{(\columnwidth - 12\tabcolsep) * \real{0.2055}}
  >{\raggedright\arraybackslash}p{(\columnwidth - 12\tabcolsep) * \real{0.1324}}
  >{\raggedright\arraybackslash}p{(\columnwidth - 12\tabcolsep) * \real{0.1324}}
  >{\raggedright\arraybackslash}p{(\columnwidth - 12\tabcolsep) * \real{0.1324}}
  >{\raggedright\arraybackslash}p{(\columnwidth - 12\tabcolsep) * \real{0.1324}}
  >{\raggedright\arraybackslash}p{(\columnwidth - 12\tabcolsep) * \real{0.1324}}
  >{\raggedright\arraybackslash}p{(\columnwidth - 12\tabcolsep) * \real{0.1325}}@{}}
\toprule
\begin{minipage}[b]{\linewidth}\raggedright
\textbf{LOD}
\end{minipage} & \begin{minipage}[b]{\linewidth}\raggedright
\textbf{Ultra\_high}
\end{minipage} & \begin{minipage}[b]{\linewidth}\raggedright
\textbf{High}
\end{minipage} & \begin{minipage}[b]{\linewidth}\raggedright
\textbf{Medium}
\end{minipage} & \begin{minipage}[b]{\linewidth}\raggedright
\textbf{Default}
\end{minipage} & \begin{minipage}[b]{\linewidth}\raggedright
\textbf{Low}
\end{minipage} & \begin{minipage}[b]{\linewidth}\raggedright
\textbf{Ultra\_low}
\end{minipage} \\
\midrule
\endhead
\textbf{Reduction Rate} & 0\% & 40\% & 65\% & 75\% & 90\% & 95\% \\
\bottomrule
\end{longtable}

This stage is introduced to simultaneously satisfy three demands.
Firstly, anatomical areas such as near the spine, joint contours, and
feature edges must be protected throughout the reduction process,
combining conservative reduction in the quality-prioritized path with
step-wise smoothing and normal recalculation to minimize shape
distortion. Secondly, since the size and noise characteristics of input
meshes can vary significantly in real-world applications, a
complexity-aware policy automatically determines retention rates and
post-processing intensity to consistently maintain output quality and
processing times amid data diversity. Lastly, different resolution
requirements coexist for the same subject, including analysis, real-time
inference, streaming, and storage; thus, standardizing the six LODs in
Table 2 and maintaining a subset hierarchy obtained by further
simplifying higher results ensures consistency in coordinates,
boundaries, and normals during resolution transitions. Consequently,
this adaptive reduction achieves accuracy, efficiency, and consistency,
enhancing the robustness of succeeding modules and the operational
efficiency of the entire pipeline.

\subsection{Estimating Skeleton based on Multiple LOD Ensembles}

The spinal skeleton is automatically estimated through a 3D human mesh
categorized into six levels of detail (LOD). The proposed skeleton
estimation module consists of the following steps:

\begin{enumerate}
\def\labelenumi{\arabic{enumi}.}
\item
  Initializing 3D joints using AI-based 2D pose landmarks;
\item
  Predicting independent skeletons from each LOD mesh;
\item
  Determining the final skeleton through median-based ensemble voting.
\end{enumerate}

The initial joint coordinates of the skeleton are derived from 2D pose
landmarks extracted from a frontal image. The frontal depth map is fed
into the pose estimation model of MediaPipe Pose to detect full body
keypoints, including the shoulders, elbows, wrists, pelvis, knees, and
ankles. Among these, key points that directly contribute to spinal
alignment and disc risk assessment are selected, including the head,
neck, upper spine, middle spine, lower spine, and both shoulder and
pelvic joints, defining the basic skeleton structure.

The initialized 3D joint coordinates are refined independently within
each of the six LOD meshes. Neighboring vertices around the joint
candidates are collected using KD-tree based nearest neighbor searches,
and the curvature and normal distribution of the local patch are
analyzed to fine-tune the joint positions to conform with the body
contour and anatomical orientation.

The final selected joint points are connected in the order of neck,
upper spine, middle spine, and lower spine to form the spinal
centerline, while the 3D curvature angles of the cervical, thoracic, and
lumbar regions are calculated through the dot product of adjacent
segment vectors. The spine is divided into three anatomical regions
using anatomical ratio-based landmarks. The vertical distance between
the C7 point, corresponding to the base of the cervical spine, and the
sacral promontory is defined as the effective spinal length. Based on
this length, the top 20\% interval is classified as the cervical region,
the 20--50\% interval as the thoracic region, and the 50--80\% interval
as the lumbar region. This proportional division method absorbs absolute
length variations due to individual height and body shape differences
while consistently reflecting the physiological positions of spinal
curvature reported in the literature. Each segment angle is also
calculated individually across the six LODs, then the value closest to
the median is adopted as the final value, minimizing estimation
deviations of spinal angles due to differences in resolution and mesh
reduction.

The skeleton is estimated by integrating the independently predicted
joint candidates from different LODs through a median-based ensemble
voting approach(Figure 7). There are differing characteristics between
the ultra low and ultra high LODs regarding quantization bias, surface
smoothing degree, and noise sensitivity. In the ultra low LOD, the mesh
resolution is significantly reduced, maintaining the overall body
contour while losing detailed structure, which decreases the accuracy of
joint position estimation. In contrast, the medium LOD strikes a balance
between structural stability and detailed shape representation, but may
show insufficient local representation at certain joints. The high and
ultra high LODs reflect fine shapes excessively, leading to high
precision; however, they are overly sensitive to noise and local
misalignments, which increases the likelihood of encountering outlier
joint angles. These characteristics suggest that simply increasing the
LOD resolution does not necessarily lead to improved estimation
accuracy. Therefore, this study adopted a median-based ensemble voting
strategy to leverage the advantages of independently predicted joint
candidates from different LODs while compensating for their
disadvantages, ultimately estimating the final skeleton.

\begin{center}
\includegraphics[width=5.37942in,height=4.23186in]{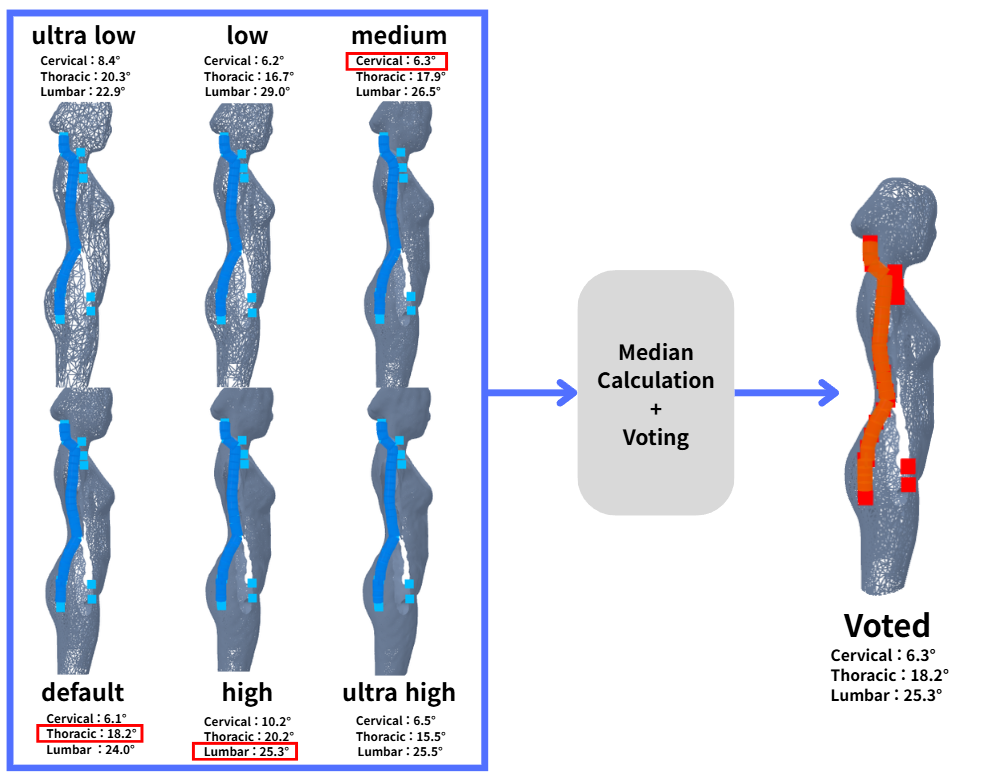}
\end{center}

\captionof{figure}{LOD-wise Joint Estimation Variation and Median-based}\label{fig:7}
Voting Result

The purpose of introducing multi-LOD ensemble is to mitigate local bias
errors and ensure consistency in joint estimation across variations in
resolution. Table 3 quantitatively evaluates how much the proposed
median-based voting strategy reduces the inter-LOD deviation compared to
average interpolation. In the cervical and lumbar regions, the
median-based voting significantly reduced the angular deviation compared
to the mean value (0.1131\textdegree{} vs. 0.9897\textdegree{} for cervical, and 0.0894\textdegree{} vs.
0.2401\textdegree{} for lumbar), while maintaining a level of consistency similar to
the mean-based approach in the thoracic region. Here, the deviation
refers to the mean absolute difference (Mean Absolute Deviation) between
each LOD candidate and the final result, with smaller values indicating
higher geometric stability (Inter-LOD Consistency) of joint estimation
across different resolutions.

\captionof{table}{Ensemble Voting: Median vs. Mean Deviation Analysis}\label{tab:3}

\begin{longtable}[]{@{}
  >{\raggedright\arraybackslash}p{(\columnwidth - 6\tabcolsep) * \real{0.2499}}
  >{\raggedright\arraybackslash}p{(\columnwidth - 6\tabcolsep) * \real{0.2501}}
  >{\raggedright\arraybackslash}p{(\columnwidth - 6\tabcolsep) * \real{0.2499}}
  >{\raggedright\arraybackslash}p{(\columnwidth - 6\tabcolsep) * \real{0.2501}}@{}}
\toprule
\begin{minipage}[b]{\linewidth}\raggedright
\textbf{Spinal Region}
\end{minipage} & \begin{minipage}[b]{\linewidth}\raggedright
\textbf{Selected LOD}
\end{minipage} & \begin{minipage}[b]{\linewidth}\raggedright
\textbf{Median Deviation}
\end{minipage} & \begin{minipage}[b]{\linewidth}\raggedright
\textbf{Mean Deviation}
\end{minipage} \\
\midrule
\endhead
Cervical & medium & 0.1131\textdegree{} & 0.9897\textdegree{} \\
Thoracic & default & 0.1915\textdegree{} & 0.1186\textdegree{} \\
Lumbar & high & 0.0894\textdegree{} & 0.2401\textdegree{} \\
\bottomrule
\end{longtable}

Average interpolation generates virtual coordinates that were not
actually observed at any LOD and reflects the inherent noise and bias
present at the LOD level. In contrast, the median-based voting approach
selects only candidates closest to the central tendency of actual
predictions, thus suppressing the influence of outliers and preserving
real anatomical structures that have been observed on the mesh surface
multiple times. Consequently, the proposed voting strategy enhances the
robustness and reliability of spinal curvature estimation by
integratively compensating for structural biases arising from different
LODs.

The reasons for adopting a multi-LOD ensemble can be summarized as
threefold. First, local quantization and smoothing biases that arise
during mesh reduction can lead to skewed joint positions when using only
a single resolution. By independently estimating joints across different
LODs and selecting the actual value closest to the central tendency,
systematic errors can be reduced through the offsetting of LOD-specific
biases. Second, uniformly sampling (50,000 points) and applying
standardized normal estimation parameters (radius = 5, max\_nn = 30)
normalize the input distribution across LODs, ensuring homogeneity in
point and normal statistics, and compensating for under-sampling or
over-sampling of certain LODs with predictions from others. Third, using
frontal image-based landmarks as a global guide helps reduce initial
uncertainties in point-based 3D estimation and encourages multiple LOD
candidates to converge under a common anatomical reference.

Choosing "the actual prediction closest to the central tendency" instead
of direct average coordinates in voting is intended to exclude
unrealistic interpolated coordinates, assuming that the internal
estimator already satisfies geometric constraints such as relative
length and angles between joints, thus maintaining only actual
mesh-based solutions. The computational complexity increases linearly
with the number of LODs; however, by processing each LOD path in
parallel or in batches, the overall computational time can be
efficiently managed. The visualization results are generated in a common
coordinate system, making it easy to compare overlays between LODs, and
when combined with the LOD hierarchy defined in Section 3.5's Adaptive
Vertex Reduction, it functions as a key module supporting real-time
inference, precise analysis, and long-distance visualization.

\subsection{Body Shape Analysis according to Spinal Angle}

This study aims to quantify the risk of cervical and lumbar discs by
utilizing metrics that reflect the body's posture and balance from the
extracted skeleton. To quantify disc risk, key body points are extracted
using MediaPipe Pose, and the angles and ratios calculated from these
coordinates are used to assess the risk of neck and back disc injuries.
To evaluate disc risk, a clear definition of body metrics is necessary.
Therefore, this study rigorously defines six variables (cervical
lordosis angle, thoracic kyphosis angle, lumbar lordosis angle, shoulder
levelness, pelvic angle, and spinal registration) that directly map to
the scoring rules of ISO 11226 (Static Posture of Workers) [35] and
RULA/REBA [36] behavior level scoring, performing posture-based disc
risk diagnostics based on these quantified results.

The justification for adopting ISO 11226 and RULA/REBA as criteria lies
in their complementary roles as an international standard (ISO) and a
practical field tool (RULA/REBA). ISO 11226 systematically defines
angular postures of the head, neck, and torso, static holding times
(holding/recovery), support status, and left-right symmetry
(tilt-rotation), allowing for the definition of posture-induced loads
acting on the lumbar and cervical discs and deriving risk indicators for
normal, caution, and risk across angle intervals [37]. RULA/REBA,
through a total of four levels of action ratings that integrate posture,
load, frequency, and coupling information, provides a tool that scores
postures such as "leaning forward," "tilting to the side," and "twisting
the torso," presenting posture risk levels while assigning situational
weighting above the absolute permissible criteria provided by
ISO[38].

Thus, according to the principles of ISO 11226 and RULA/REBA, if the
angle of "leaning forward" for the neck and torso exceeds approximately
20\textdegree{}, or if there is a pronounced "tilt to the side" or "twisting"
(asymmetry between shoulders and pelvis) in the frontal and rear views
and an increase in global registration (SVA), the action level is
escalated, and posture ratings are refined[39]. This dual assessment
structure allows for the objective evaluation of neck and lumbar disc
risks using only 2D key points extracted from frontal, rear, and side
images, applying standards-compliant rules and providing grounded
indicators for normal/caution/risk across angle intervals.

ISO 11226 describes the posture-induced mechanical loads on the neck and
lumbar region based on angles (head/neck, torso), static holding times,
support status, and left-right symmetry, while RULA/REBA scores forward
flexion, lateral tilt, torso twisting, as well as load, frequency of
repetition, and object handling status (coupling) to indicate the
urgency of interventions. Based on these definitions, this study selects
six variables necessary for evaluating neck and lumbar discs: Cervical
Lordosis Angle, Thoracic Kyphosis Angle, Lumbar Lordosis Angle, Shoulder
Angle, Pelvic Angle, and Sagittal Vertical Axis. Based on the defined
variables, calculations for each variable are based on the MediaPipe
keypoint. Figure 8 shows the key points of the body that are utilized in
the variables. The calculation formula and definition of the variables
are set as follows.

\begin{center}
\includegraphics[width=3.14618in,height=2.81669in]{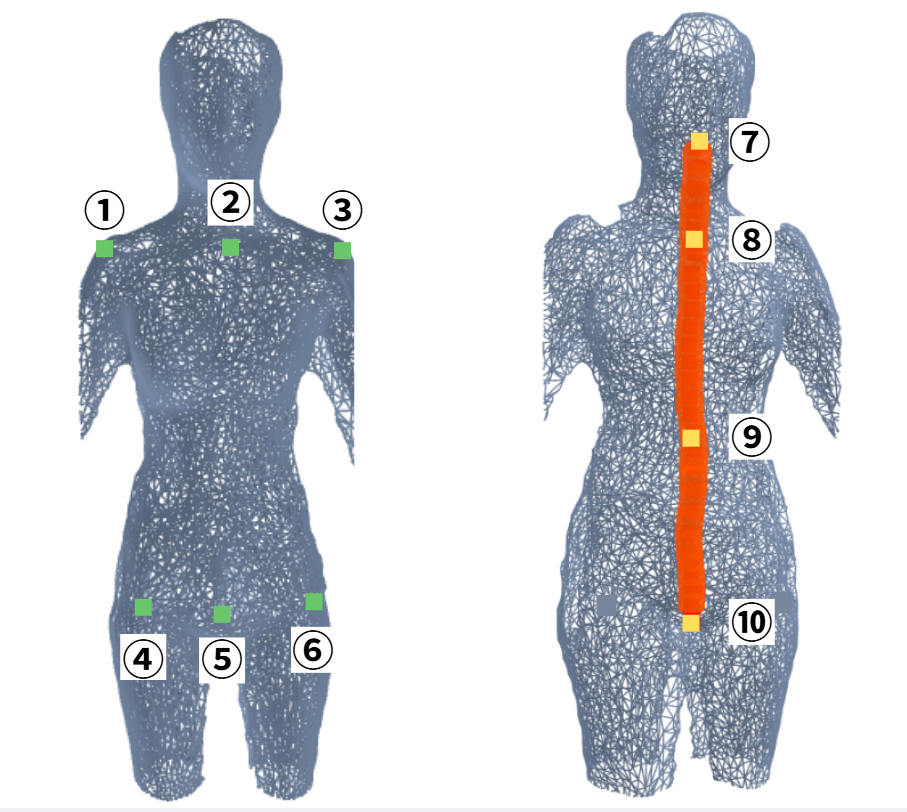}
\end{center}

\captionof{figure}{Anatomical locations of keypoint landmarks on the 3D body model(1 : right shoulder(\({sh}_{R}\)), 2 : shoulder
center(\(m_{sh}\)), 3: left shoulder(\({sh}_{L}\)), 4: right
hip(\({Hip}_{R}\)), 5: hip center(\(m_{hip}\)), 6: left
hip(\({Hip}_{L}\)), 7: neck, 8 : upper spine, 9: middle spine, 10: lower
spine}

\begin{itemize}
\item
  Cervical Lordosis Angle (\(\theta_{N}\))

  The cervical lordosis angle is approximated using neck flexion,
  defined as the difference between the trunk flexion angle ($\alpha$) and the
  head flexion angle ($\beta$). The values of $\alpha$ and $\beta$ are given by equations
  (1) and (2), respectively, and the cervical lordosis angle is derived
  using their difference as shown in equation (3). Therefore, the larger
  the angle, the more pronounced the loss of cervical lordosis and
  forward head posture become, which directly maps to the neck flexion
  item in ISO and the neck forward angle score in RULA [40].
  \(\overrightarrow{v\ }\) represents the vertical reference vector, and
  as \(\theta_{N}\) increases, the cervical lordosis decreases, which
  directly corresponds to the neck flexion item in ISO and the neck
  forward angle score in RULA.
\end{itemize}

\begin{longtable}[]{@{}
  >{\raggedright\arraybackslash}p{(\columnwidth - 2\tabcolsep) * \real{0.9452}}
  >{\raggedright\arraybackslash}p{(\columnwidth - 2\tabcolsep) * \real{0.0548}}@{}}
\toprule
\begin{minipage}[b]{\linewidth}\raggedright
\[\alpha = \angle(\ \overrightarrow{m_{hip},m_{sh}\ }\ ,\overrightarrow{v\ }\ )\]
\end{minipage} & \begin{minipage}[b]{\linewidth}\raggedright
(1)
\end{minipage} \\
\midrule
\endhead
\(\beta = \angle(\overrightarrow{Ear,EyeOuter\ }\ ,\overrightarrow{v\ })\)
& (2) \\
\(\theta_{N} = \beta - \alpha\) & (3) \\
\bottomrule
\end{longtable}

\begin{itemize}
\item
  Thoracic Kyphosis Angle (\(\theta_{CT}\))

  The thoracic kyphosis angle is approximated using the upper and lower
  trunk segment angles. It is calculated as the angle between the vector
  directed from the midpoint of the shoulders to the ear and the vector
  directed from the midpoint of the shoulders to the hip joint midpoint,
  as shown in equation (4). A larger value is interpreted as a tendency
  toward hyperkyphotic. As \(\theta_{CT}\) increases, the thoracic
  kyphosis deepens, bringing the posture closer to a kyphotic posture
  [41].
\end{itemize}

\begin{longtable}[]{@{}
  >{\raggedright\arraybackslash}p{(\columnwidth - 2\tabcolsep) * \real{0.9452}}
  >{\raggedright\arraybackslash}p{(\columnwidth - 2\tabcolsep) * \real{0.0548}}@{}}
\toprule
\begin{minipage}[b]{\linewidth}\raggedright
\[\theta_{CT} = \angle(\ \overrightarrow{m_{sh}\ ,Ear},\ \overrightarrow{m_{sh}\ ,m_{hip}}\ )\]
\end{minipage} & \begin{minipage}[b]{\linewidth}\raggedright
(4)
\end{minipage} \\
\midrule
\endhead
\bottomrule
\end{longtable}

\begin{itemize}
\item
  Lumbar Lordosis Angle (\(\theta_{TL}\))

  The lumbar lordosis angle is approximated using the
  thoracolumbar-pelvic segment angle. It is calculated as the angle
  between the vector directed from the hip joint midpoint to the
  shoulder midpoint and the vector directed from the hip joint midpoint
  to the knee midpoint, as shown in equation (5). A smaller value of
  \(\theta_{TL}\) indicates a tendency toward loss of lordosis or a flat
  back posture [42].
\end{itemize}

\begin{longtable}[]{@{}
  >{\raggedright\arraybackslash}p{(\columnwidth - 2\tabcolsep) * \real{0.9452}}
  >{\raggedright\arraybackslash}p{(\columnwidth - 2\tabcolsep) * \real{0.0548}}@{}}
\toprule
\begin{minipage}[b]{\linewidth}\raggedright
\[\theta_{TL} = \angle(\overrightarrow{m_{hip}\ ,m_{sh}},\ \overrightarrow{m_{hip}\ ,m_{knee}})\]
\end{minipage} & \begin{minipage}[b]{\linewidth}\raggedright
(5)
\end{minipage} \\
\midrule
\endhead
\bottomrule
\end{longtable}

\begin{itemize}
\item
  Shoulder Angle (\(\theta_{sh}\))

  Shoulder levelness is calculated as the angle between the line
  connecting the left and right shoulders and the horizontal line, as
  shown in equation (6). The indicators for the shoulders and pelvis
  (equation (6) and (7)) quantify coronal plane symmetry violations
  (lateral bending/twisting) and are used as scoring items for lateral
  bending and twisting in RULA/REBA [43].
\end{itemize}

\begin{longtable}[]{@{}
  >{\raggedright\arraybackslash}p{(\columnwidth - 2\tabcolsep) * \real{0.9452}}
  >{\raggedright\arraybackslash}p{(\columnwidth - 2\tabcolsep) * \real{0.0548}}@{}}
\toprule
\begin{minipage}[b]{\linewidth}\raggedright
\[\theta_{sh} = \angle(\overrightarrow{{sh}_{L}\ ,{sh}_{R}},\ h)\]
\end{minipage} & \begin{minipage}[b]{\linewidth}\raggedright
(6)
\end{minipage} \\
\midrule
\endhead
\bottomrule
\end{longtable}

\begin{itemize}
\item
  Pelvic Angle (\(\theta_{pelvis}\))

  Pelvic tilt (obliquity) is calculated as the angle between the line
  connecting the left and right hip joints and the horizontal line, as
  shown in equation (7).
\end{itemize}

\begin{longtable}[]{@{}
  >{\raggedright\arraybackslash}p{(\columnwidth - 2\tabcolsep) * \real{0.9452}}
  >{\raggedright\arraybackslash}p{(\columnwidth - 2\tabcolsep) * \real{0.0548}}@{}}
\toprule
\begin{minipage}[b]{\linewidth}\raggedright
\[\theta_{pelvis} = \angle(\overrightarrow{{Hip}_{L}\ ,{Hip}_{R}},\ h)\]
\end{minipage} & \begin{minipage}[b]{\linewidth}\raggedright
(7)
\end{minipage} \\
\midrule
\endhead
\bottomrule
\end{longtable}

\begin{itemize}
\item
  Sagittal Vertical Axis (SVA)

  The spinal alignment is measured using a photo-based Sagittal Vertical
  Axis (SVA). It is calculated by normalizing the difference between the
  horizontal coordinate of the head reference point and the horizontal
  coordinate of the hip joint midpoint, scaled by the total body length,
  as shown in equation (8) [44]. In this case, a larger SVA value
  indicates a greater deviation from vertical alignment of the entire
  body.
\end{itemize}

\begin{longtable}[]{@{}
  >{\raggedright\arraybackslash}p{(\columnwidth - 2\tabcolsep) * \real{0.9452}}
  >{\raggedright\arraybackslash}p{(\columnwidth - 2\tabcolsep) * \real{0.0548}}@{}}
\toprule
\begin{minipage}[b]{\linewidth}\raggedright
\[SVA = \ \frac{x_{Ear} - {\ x}_{m_{hip}}}{H_{pix}}\  \times \ H_{real}\ \]
\end{minipage} & \begin{minipage}[b]{\linewidth}\raggedright
(8)
\end{minipage} \\
\midrule
\endhead
\bottomrule
\end{longtable}

This calculation procedure allows for the consistent mapping of the
criteria of ``angles, static holding, and symmetry'' from ISO 11226 and
the principles of ``situational weighting (forward bending, lateral
tilting, twisting, and load/frequency/coupling)'' from RULA/REBA, using
only the 2D key points provided by MediaPipe. Each variable can then be
combined with the defined angle ranges (normal/caution/risk) to serve as
input indicators for diagnosing the risk of cervical and lumbar discs.
Table 4 summarizes the normal/caution/risk ranges for spinal angles as
defined in this study. To enable a structured interpretation of the extracted spinal indicators, this study adopts posture angle ranges derived from the principles of ISO 11226 and RULA/REBA. Within the context of depth-based, non-contact posture analysis, these ranges are utilized to de-scribe relative postural tendencies and to facilitate the stratification of posture conditions into normal, caution, and risk categories. Given the inherent measurement variability and quantization uncertainty associated with depth sensors, the use of stratified angle ranges is intended to enhance robustness and interpretability at the posture risk level rather than precise angle estimation.

Accordingly, the criteria summarized in Table 4 provide a consistent reference framework for ergonomic posture screening and longitudinal monitoring, rather than ab-solute clinical decision boundaries. This represents a redefinition of the risk level ranges specified in ISO 11226 and RULA/REBA within the spinal coordinate system adopted in this experiment. Accordingly, based on these posture assessment principles, the following table presents screening-level angle ranges adapted for depth-based spinal posture analy-sis. 

\captionof{table}{Criteria for risk of disks through extracted indicator}\label{tab:4}
value

\begin{longtable}[]{@{}
  >{\raggedright\arraybackslash}p{(\columnwidth - 8\tabcolsep) * \real{0.2889}}
  >{\raggedright\arraybackslash}p{(\columnwidth - 8\tabcolsep) * \real{0.0733}}
  >{\raggedright\arraybackslash}p{(\columnwidth - 8\tabcolsep) * \real{0.2125}}
  >{\raggedright\arraybackslash}p{(\columnwidth - 8\tabcolsep) * \real{0.2126}}
  >{\raggedright\arraybackslash}p{(\columnwidth - 8\tabcolsep) * \real{0.2126}}@{}}
\toprule
\begin{minipage}[b]{\linewidth}\raggedright
\textbf{Cervical and lumbar disc evaluation variables}
\end{minipage} & \begin{minipage}[b]{\linewidth}\raggedright
\textbf{View}
\end{minipage} & \begin{minipage}[b]{\linewidth}\raggedright
\textbf{Normal}
\end{minipage} & \begin{minipage}[b]{\linewidth}\raggedright
\textbf{Caution}
\end{minipage} & \begin{minipage}[b]{\linewidth}\raggedright
\textbf{Danger}
\end{minipage} \\
\midrule
\endhead
Cervical Lordosis Angle & Side & 20 \textasciitilde{} 35\textdegree{} & 10
\textasciitilde{} 20\textdegree{} or 35 \textasciitilde{} 45\textdegree{} & \textless{} 10\textdegree{} or
\textgreater{} 45\textdegree{} \\
Thoracic Kyphosis Angle & Side & 20 \textasciitilde{} 40\textdegree{} & 15
\textasciitilde{} 20\textdegree{} or 40 \textasciitilde{} 55\textdegree{} & \textless{} 15\textdegree{} or
\textgreater{} 55\textdegree{} \\
Lumbar Lordosis Angle & Front & 40 \textasciitilde{} 60\textdegree{} & 30
\textasciitilde{} 40\textdegree{} or 60 \textasciitilde{} 70\textdegree{} & \textless{} 30\textdegree{} or
\textgreater{} 70\textdegree{} \\
Shoulder Angle & Front & $\leq$ 2\textdegree{} & 2 \textasciitilde{} 10\textdegree{} & \textgreater{}
10\textdegree{} \\
Pelvic Angle & Side & $\leq$ 3\textdegree{} & 3 \textasciitilde{} 10\textdegree{} & \textgreater{}
10\textdegree{} \\
Sagittal Vertical Axis & Side & \textless{} 4 cm & 4 \textasciitilde{} 6
cm & \textgreater{} 6 cm \\
\bottomrule
\end{longtable}

\section{Experimental Results}

\subsection{Experimental Environment}

The programming language used is Python version 3.10.18, and the main
libraries include OpenCV for image processing, TorchGeometry, NumPy
($\geq$1.21.0) and SciPy ($\geq$1.7.0) for numerical calculations, Pillow ($\geq$8.3.0)
for image input/output processing, Open3D ($\geq$0.15.0) for 3D data
processing, and MediaPipe ($\geq$0.8.9) for human pose recognition. All experiments were executed on CPU without GPU acceleration. The hardware configu-ration consists of an Intel Core i5-14600 CPU, 31GB of system RAM, and approximately 80GB of disk space used for experiments. The input data
comprises depth maps with a resolution of 512$\times$512, sampled at every 4
pixels (step = 4), with a depth scale adjusted to 100. A 3D integrated
model was generated by aligning point clouds from the left, right, and
rear views based on the frontal view. The registration process consists
of two stages: FPFH-based RANSAC global registration (Global
Registration) and Point-to-Plane ICP-based fine registration (Local
Refinement). The registration parameters for each view direction are
shown in Table 5.

To ensure bilateral consistency and reproducibility, we apply identical registration parameters to the left and right side views by explicitly reflecting the anatomical bilateral symmetry of the human body. Since the left/right side views exhibit comparable geometric complexity and similar overlap/occlusion patterns, using a standardized RANSAC budg-et avoids view-dependent bias in hypothesis sampling and yields more consistent align-ment outcomes. Accordingly, we set the RANSAC iteration count to 50,000 for both left and right views. In contrast, the back view is integrated after the side views, benefiting from the already aligned lateral structure, which effectively reduces the transformation search space. Therefore, 30,000 iterations are used for the back view to maintain efficiency while preserving robust alignment performance. This symmetric parameter design im-proves bilateral consistency and enhances the stability and reproducibility of the over-all pipeline.

\captionof{table}{Registration Parameters by View Direction}\label{tab:5}

\begin{longtable}[]{@{}
  >{\raggedright\arraybackslash}p{(\columnwidth - 10\tabcolsep) * \real{0.11}}
  >{\raggedright\arraybackslash}p{(\columnwidth - 10\tabcolsep) * \real{0.11}}
  >{\raggedright\arraybackslash}p{(\columnwidth - 10\tabcolsep) * \real{0.25}}
  >{\raggedright\arraybackslash}p{(\columnwidth - 10\tabcolsep) * \real{0.14}}
  >{\raggedright\arraybackslash}p{(\columnwidth - 10\tabcolsep) * \real{0.14}}
  >{\raggedright\arraybackslash}p{(\columnwidth - 10\tabcolsep) * \real{0.14}}@{}}
\toprule
\begin{minipage}[b]{\linewidth}\raggedright
\textbf{View of Point Cloud}
\end{minipage} & \begin{minipage}[b]{\linewidth}\raggedright
\textbf{Coarse\\
Voxel}\strut
\end{minipage} & \begin{minipage}[b]{\linewidth}\raggedright
\textbf{Voxel\\
List(ICP)}\strut
\end{minipage} & \begin{minipage}[b]{\linewidth}\raggedright
\textbf{RANSAC\\
Iterations}\strut
\end{minipage} & \begin{minipage}[b]{\linewidth}\raggedright
\textbf{Rotation\\
Allowed}\strut
\end{minipage} & \begin{minipage}[b]{\linewidth}\raggedright
\textbf{Small\\
Rotation}\strut
\end{minipage} \\
\midrule
\endhead
Left & 4.0 & [ 15.0, 8.0, 4.0, 2.0] & 50,000 & O & O ($\pm$2\textdegree{}) \\
Right & 4.0 & [ 15.0, 8.0, 4.0, 2.0] & 50,000 & O & O ($\pm$2\textdegree{}) \\
Back & 5.0 & [ 25.0, 12.0, 6.0] & 30,000 & O & O ($\pm$2\textdegree{}) \\
\bottomrule
\end{longtable}

During the FPFH + RANSAC alignment, the maximum correspondence distance
was set to 2.5 times the voxel size, with a minimum required number of
correspondences set to 3 and a confidence level relaxed to 0.95. In the
ICP stage, the correspondence distances for each scale were set to 3.0
times the voxel size, with the number of iterations configured as
follows: [150, 200, 250, 300]. For normal estimation, the radius was
set to 2.0 times the voxel size, and the maximum number of nearest
neighbors (max\_nn) was set to 30. Additionally, to achieve fine
registration with minimal rotation, Small Rotation ICP was applied, with
maximum rotation angles for each scale set to [2.0\textdegree{}, 1.0\textdegree{}, 0.5\textdegree{}] and
iteration counts as [30, 30, 60]. For regions with restricted
rotation, Translation-only ICP was performed, applying the following
parameters: iteration counts of [20, 20, 30] and a correspondence
distance coefficient of 1.5.

\subsection{Results}

Figure 9 visually illustrates the intermediate outputs generated at each
stage according to the proposed system's overall pipeline. Initially,
unnecessary background is removed from the input four-direction depth
images, creating independent point clouds for each viewpoint. Thanks to
the mask-based preprocessing, the human outline is clearly delineated,
and the fine surface shapes of regions such as the hands, feet, and head
are relatively stably extracted.

\begin{center}
\includegraphics[width=6.44736in]{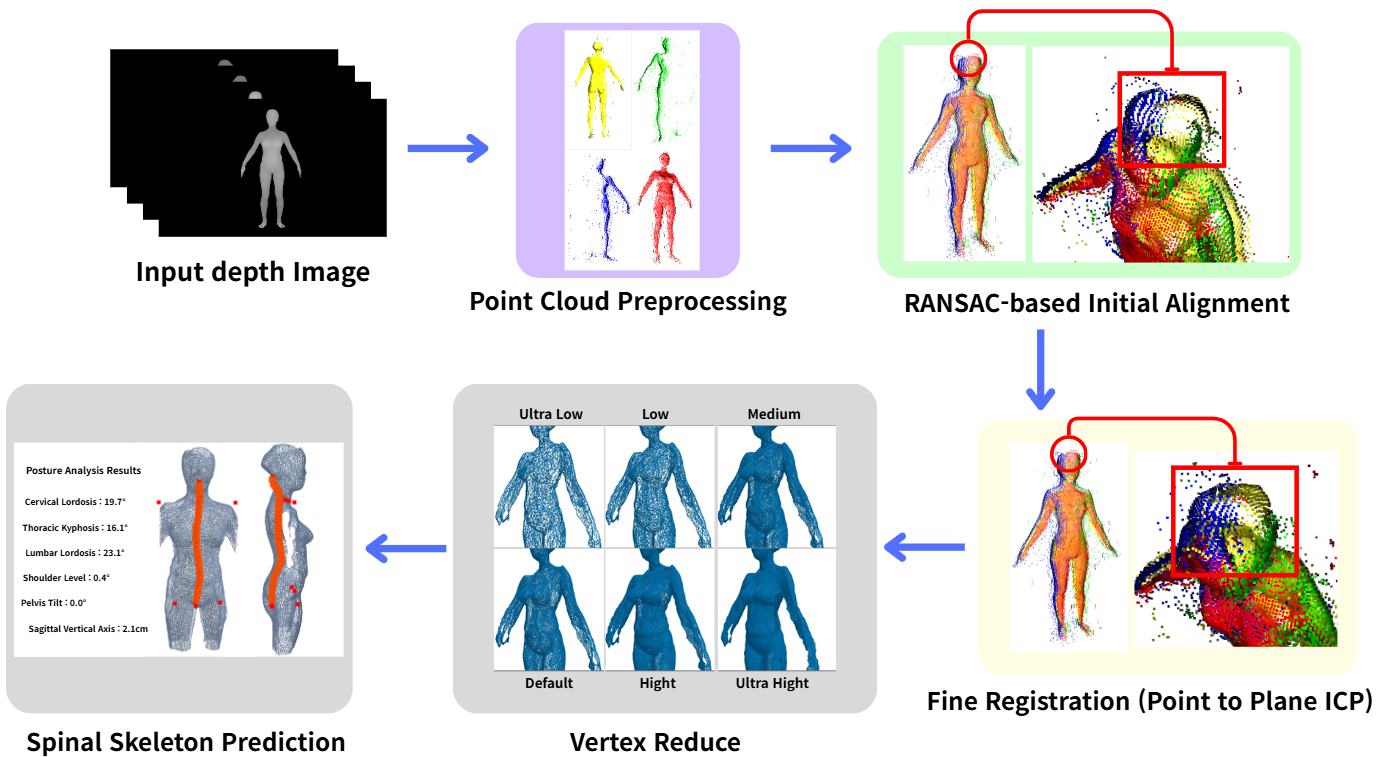}
\end{center}

\captionof{figure}{Intermediate Output Results by Pipeline Sequence.}\label{fig:9}

In the RANSAC-based initial registration stage, a global registration
initial value is formed among the point clouds of the four viewpoints
based on FPFH feature matching. As shown in the figure, even with this
initial alignment, the relationships between the body center axis and
limb positions are largely consistent, and the overall body contour
begins to integrate into a single structure. However, the head region
exhibits somewhat unstable alignment at this initial stage due to
complex geometric patterns caused by hair, contour variations, and the
specular reflection in the depth images. This instability arises even
though FPFH uses local normal-based features, as the surface
directionality of microstructures like hair is inconsistent.

In the Fine Registration stage, the Point-to-Plane ICP algorithm is
applied to optimize the local correspondences among the point clouds
more precisely. At this stage, the registration errors in complex
curvature areas such as the head, shoulders, waist, and both arms are
significantly reduced, and the slight displacements experienced in the
initial stage are gradually corrected. Particularly, the registration
quality improves substantially in the top and temporal regions of the
head, ultimately confirming that the entire body is seamlessly
integrated into a coherent 3D shape.

This visual comparison demonstrates that the step-wise design of the
system is effectively practical, with FPFH-based initial registration
rapidly stabilizing the overall structure and ICP fine registration
meticulously refining the detailed geometry.

Figure 10 shows the reconstructed spinal centerline (in orange) and
shoulder and pelvic reference lines by the proposed system for both male and female domains. The bottom row additionally includes subjects with obese body types, where the overall registration remains stable and the extracted reference lines re-main visually coherent. In both the frontal and side views, the spinal
centerline is naturally connected along the body's central axis, and the
symmetry of the shoulders and pelvis is stably maintained. Notably, in
female subjects, the lumbar lordosis appears more pronounced due to body
shape characteristics, while in male subjects, the curvature of the
thoracic kyphosis is relatively gentle. These visual results demonstrate
that the proposed system can reliably reconstruct the major anatomical
axes (spine, pelvis, shoulders) of the human body, even with incomplete
depth information from a single viewpoint.

Additionally, each visualized model presents detailed angular
measurements of the spine, allowing for quantitative comparisons of
cervical lordosis, thoracic kyphosis, lumbar lordosis, shoulder
levelness, pelvic tilt, and sagittal vertical axis (SVA) for each
subject. These measurements were evaluated against the normal body shape
standards in \textless Table 4\textgreater{} to assess postural
registration.

The results show that most subjects fell within the ranges of cervical
lordosis of 20--35\textdegree{}, thoracic kyphosis of 20--40\textdegree{}, and lumbar lordosis
of 40--60\textdegree{}, classifying them as normal body shapes. However, some female
subjects exhibited lumbar lordosis angles above 40\textdegree{}, indicating a
tendency towards lumbar hyperlordosis, while some male subjects showed
thoracic kyphosis angles below 10\textdegree{}, presenting a flat thoracic posture.
Despite these variations, all subjects maintained shoulder levelness and
pelvic tilt within 2\textdegree{} or less, indicating good left-right balance, with
an average SVA of 1.1 cm, which is within the normal standard of
\textless{} 4 cm.

These findings suggest that the proposed system effectively utilizes
four-direction depth maps to robustly address surface noise and
viewpoint variations while maintaining spatial consistency in the spinal
centerline, pelvic axis, and shoulder line, regardless of gender or body
shape differences. Furthermore, the multi-LOD (Level of Detail)
ensemble-based skeleton voting technique confirms that the spinal
centerline is continuously estimated without interruption, even in
low-resolution meshes. Consequently, the method demonstrated in this
study provides a universal 3D posture analysis framework capable of
quantitatively analyzing and visually representing differences in spinal
curvature and trunk registration based on body shape characteristics
between genders.

\includegraphics[width=5.4in]{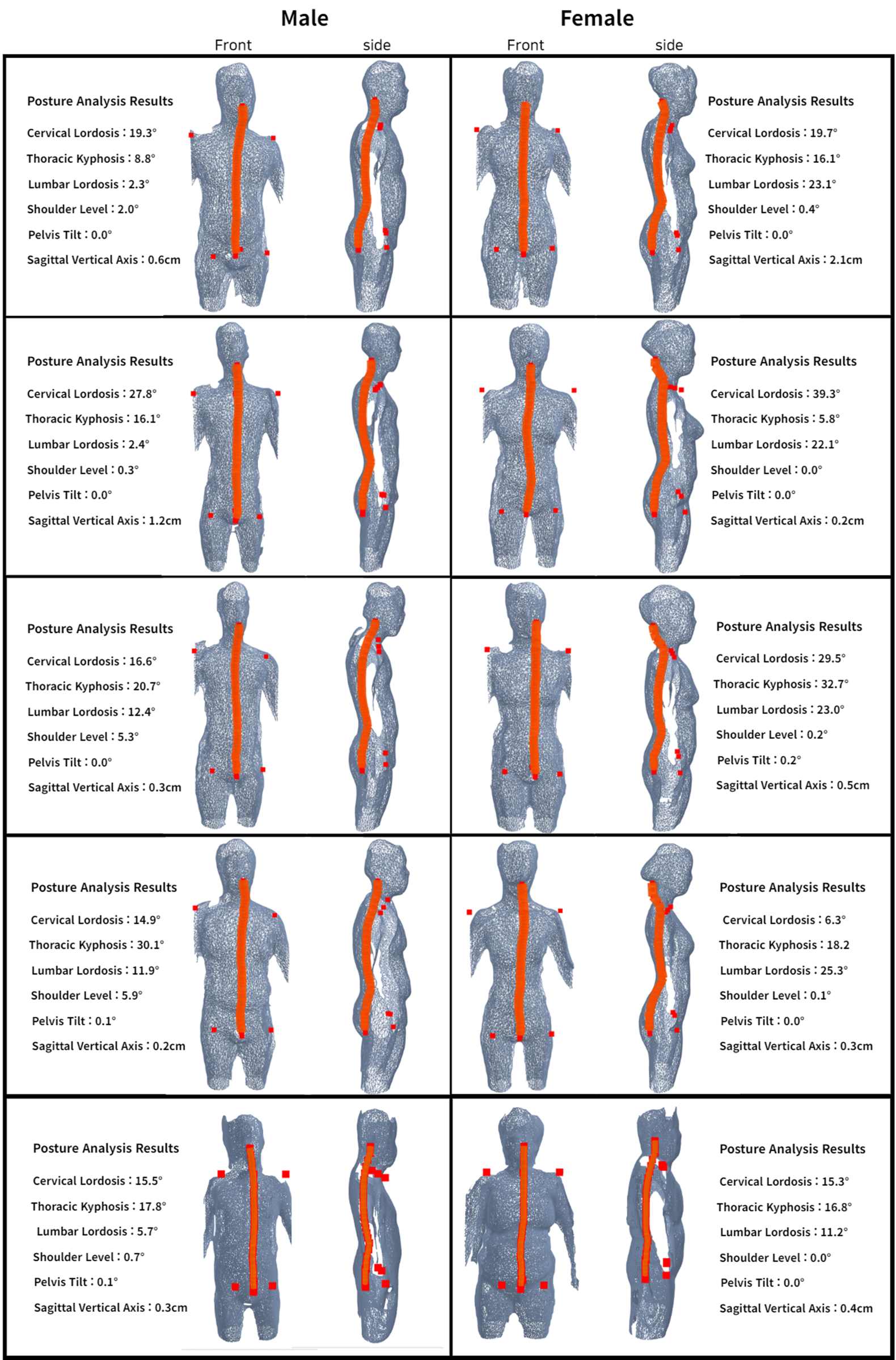}

\captionof{figure}{Results of proposed method for two domains (male/female), including subjects with obese body types (last row): The orange lines represent the spinal, pelvic, and shoulder skeletal lines}

\subsection{Statistical Analysis}
To assess whether the performance differences between Point-to-Point ICP and Point-to-Plane ICP are statistically significant, we performed paired statistical tests on the fine registration outcomes. For each multi-view reconstruction sample (N = 393), we measured ICP fitness and RMSE after local refinement under identical global initialization (RANSAC-FPFH) and correspondence update settings. RMSE and fitness follow the defi-nitions described in Section 3.4.

Normality was evaluated using the Shapiro–Wilk test. Both fitness and RMSE dis-tributions significantly deviated from normality (p < 0.001 for all cases); therefore, we adopted a non-parametric paired comparison. For RMSE, 11 paired measurements con-tained non-finite values (infinite RMSE) due to non-convergent registrations; these pairs were excluded, resulting in 376 valid paired samples for RMSE analysis.

The Wilcoxon signed-rank test showed that the fitness difference was not statistically significant (W = 14,727; p = 9.91e-2), although Point-to-Plane ICP achieved a marginally higher mean fitness (0.6610 vs. 0.6586; +0.37\%). In contrast, the RMSE difference was sta-tistically significant on the valid paired set (W = 26,774; p = 5.88e-4, p < 0.001), indicating that Point-to-Plane ICP yields a small but consistent reduction in RMSE across samples (+0.75\% improvement; lower is better) over the 376 valid pairs.

We additionally report effect sizes based on Cohen’s d computed from paired differ-ences, which were small for both metrics (fitness: d = 0.0138; RMSE: d = 0.0188). Taken to-gether, despite the marginal effect sizes, Point-to-Plane ICP demonstrates a statistically significant and consistent improvement in RMSE (with a slight gain in mean fitness), and therefore we adopt Point-to-Plane ICP as the fine registration step in the proposed pipe-line. 

\subsection{Processing Time}
To address execution-time considerations for clinical deployment, we report the wall-clock processing time of the proposed pipeline.

All timing measurements were obtained on the same workstation described in Sec-tion 4.1 using Python wall-clock timing. In a full batch run over 131 datasets, the total processing time was 98.5 min (5,910 s), corresponding to 45.1 ± 8.4 s per dataset (min: 28.3 s; max: 67.8 s). This yields a throughput of 79.9 datasets/hour, with an overall processing efficiency of 95.2\% as reported by our runtime logger. This runtime includes point cloud generation from multi-view depth images, preprocessing, RANSAC-FPFH global align-ment, Point-to-Plane ICP refinement, mesh construction, LOD generation, and LOD-ensemble skeleton extraction.

The current implementation is not designed for real-time interaction (e.g., per-frame feedback) and is intended for offline reconstruction and clinical screening workflows. Given the measured throughput (~80 datasets/hour), the proposed system supports batch-mode clinical pipelines, while real-time applications would require further optimi-zation and/or GPU-accelerated implementations. Because the pipeline is modular, users can reduce execution time by lowering the number of LODs evaluated or by relaxing glob-al-registration settings (e.g., fewer RANSAC hypotheses) when a faster but less robust mode is acceptable.

\subsection{Limitations}
As illustrated in the last row of Figure 10, the proposed pipeline can still achieve sta-ble point-cloud registration even for atypical body types such as obesity. However, it is important to note that the spinal curvature angles reported in this study are not derived from direct observation of the vertebrae, but are instead estimated from the reconstructed external body surface obtained from depth measurements. Consequently, for subjects with substantial soft-tissue thickness or atypical body proportions, the clinical accuracy of curvature-related metrics (e.g., cervical/lumbar lordosis and thoracic kyphosis) cannot be guaranteed. In such cases, the reported angles should be interpreted as screening-level in-dicators rather than definitive diagnostic measurements.

Furthermore, dense hair and loose or thick clothing may introduce depth discontinu-ities and irregular surface microstructures, which can negatively affect surface-normal es-timation and correspondence matching, particularly around the head and shoulder re-gions. To mitigate these effects, standardized acquisition conditions are recommended, such as minimizing hair-induced occlusions and using clothing that preserves the body silhouette.

The proposed approach also assumes quasi-static capture with a consistent neutral pose across the four directional views. Large pose variations, self-occlusions, or inter-view motion reduce the overlapping surface area and can lead to non-convergent registration or inconsistent skeletal line extraction. Therefore, the method may be less reliable for com-plex poses or dynamic movements, and stable acquisition conditions with limited motion are preferred.

Finally, robust performance requires sufficiently dense and well-calibrated depth measurements. Increased sensor noise, missing returns, or low spatial resolution can propagate to point-cloud artifacts, degrading both global and local registration and ulti-mately affecting the reliability of the extracted reference lines and angle estimates. Hence, the proposed pipeline is contingent on a minimum level of depth-sensor quality and sta-ble calibration for consistent results. 

Additionally, direct quantitative comparisons with deep learning-based methods   (e.g., PointNetLK [28], DCP [29]) on standard benchmarks like SMPL were not conducted in thisstudy. This decision stems from fundamental mismatches between our medical depth data—characterized by four-directional fixed-capture depth maps with specific scale and structural distributions—and the large-scale RGB-D/general human scan da-tasets used to train these models. Applying pre-trained learning-based models to our medical data without extensive fine-tuning or domain adaptation could lead to mislead-ing performance compar-isons. **Instead, we prioritized validation of geometric registra-tion accuracy and stability under clinical acquisition constraints, which aligns with the primary goal of devel oping a training-free, interpretable pipeline for healthcare applica-tions. 

Regarding global registration, methods like TEASER++, Faster Global Registration, and Go-ICP—while offering strong theoretical guarantees—were not employed due to their computational overhead, which is prohibitive for real-time clinical workflows. These ap-proaches incur significantly higher processing times (often 10-50x slower) and require ex-tensive preprocessing, making them less suitable for point-of-care applications. Our RAN-SAC-FPFH pipeline achieves comparable robustness with sub-second execution on standard hardware, optimally balancing speed and accuracy under medical acquisition con-straints.

\section{Conclusion}

This study proposed a system for reconstructing a three-dimensional
model of the human body by aligning Depth Maps acquired from four
directions: front, back, left, and right. The proposed approach involved
(1) global registration using RANSAC (Random Sample Consensus) and FPFH
(Fast Point Feature Histogram) and (2) a fine registration process
through the Point-to-Plane ICP (Iterative Closest Point) algorithm,
enabling stable and precise restoration of the body's central structure
without the need for separate training data or complex deep learning
neural network models. Additionally, (3) adaptive LOD generation that
preserves shape fidelity and multi-LOD ensemble skeleton estimation
provided robustness against resolution bias and voids.

Experimental results demonstrated the capability to combine multi-view
Depth Map data to create a robust 3D human model, with homogeneous
integration of stable front, back, left, and right shapes resistant to
noise. The generated model was utilized for estimating the spinal
centerline, allowing intuitive morphological analysis of the
registration status of the body's central axis.

A key contribution of this study is the presentation of a 3D analysis
framework that can robustly reconstruct human structures purely through
geometric registration and shape analysis without complex deep learning
networks or extensive additional training data. Furthermore, the
adaptive mesh simplification and multi-LOD-based skeleton voting reduced
resolution bias while enhancing robustness against voids and noise. This
approach maximally avoids fundamental constraints in medical imaging
environments, such as data sparsity, privacy concerns, and challenges in
generalization, while maintaining the interpretability of each
algorithmic step.

Future research could focus on improving the quality of 3D human model
reconstruction from this system. Currently, only Vertex Reduction and
Mesh Smoothing are performed on aligned models for spinal centerline
estimation. However, introducing hole-filling techniques could
facilitate continuous and natural surface restoration in missing areas,
leading to more complete and detailed 3D model reconstructions.

To further extend this study clinically, utilizing specialized medical
data such as X-ray or CT scans could enhance the geometric accuracy of
the estimated spinal centerline by simultaneously reflecting both
internal skeletal structures and external shapes. This may enable the
techniques developed in this research to evolve into a medical
anatomical 3D registration platform applicable for clinical diagnosis,
surgical planning, and rehabilitation prognosis evaluation, extending
beyond simple spinal estimation.

In addition, To further extend this study toward clinical research contexts, future work may incorporate specialized medical imaging data such as X-ray or CT scans. Such integration could enable a more detailed investigation of the relationship between external body geometry and internal skeletal structures. In the present study, the primary focus is placed on validating a depth-based geometric reconstruction framework, while clinical validation and diagnostic applications are left for future investigation.  

Consistent with prior studies adopting ISO 11226 and RULA/REBA for vision- or depth-based posture analysis, the posture risk categories defined in this work are intended to support structured interpretation of posture tendencies in non-contact settings. Future studies that further align posture risk criteria with clinically grounded standards may enhance the precision and applicability of depth-based posture risk assessment.

\textbf{Author Contributions:} S. Kim, H.J. Lee, J. Lee and T. Lee
participated in all phases and contributed equally to this work. Their
contributions to this paper are as follows. S. Kim contributed to the
methodology, software, formal analysis. H.J. Lee and J. Lee contributed
to the investigation, resources, data curation. C. Kim contributed to
the data preparation, original draft preparation, visualization, and
project administration. The contributions of T.L. were reviewing and
editing of the writing, supervision, and project administration. Also,
both contributed to the conceptualization and validation of the paper.
All authors have read and agreed to the published version of the
manuscript.

\textbf{Funding:} This research was supported by Technology Innovation
Program (RS-2024-00507228, Development of process upgrade technology for
AI self-manufacturing in the cement industry) funded By the Ministry of
Trade, Industry \& Energy(MOTIE, Korea).

\textbf{Acknowledgments:} This study was supported by Kangwon National
University.

\textbf{Conflicts of Interest:} The authors declare no conflicts of
interest.

\end{document}